\PassOptionsToPackage{dvipsnames}{xcolor}
\documentclass[sigconf]{acmart}

\usepackage{bm}
\usepackage{multirow}
\usepackage{xcolor}
\usepackage{colortbl}
\usepackage{makecell}
\usepackage{booktabs}
\usepackage{subcaption}

\usepackage{amsmath,amssymb}
\usepackage{algorithm}      
\usepackage{algpseudocode}

\usepackage[dvipsnames]{xcolor}
\usepackage{pifont}

\newcommand{\cmark}{\textcolor{ForestGreen}{\ding{51}}}
\newcommand{\xmark}{\textcolor{red}{\ding{55}}}

\def\ours{FLaG}
\definecolor{mygray}{gray}{.93}

\newtheorem{assumption}{Assumption}

\AtBeginDocument{%
  }

\setcopyright{acmlicensed}
\copyrightyear{2026}
\acmYear{2026}
\acmDOI{10.1145/3770855.3818137}
\acmConference[KDD'26]{the 32nd ACM SIGKDD Conference on Knowledge Discovery and Data Mining}{Augest 09--13,
  2026}{Jeju, Korea}
\acmISBN{979-8-4007-2258-5/2026/08}

\begin{document}

\title{FLaG: Fine-Grained Latent Grouping for Hallucination Detection}

\author{Wentao Ye}
\authornote{Both authors contributed equally to this research.}
\email{yewt01@zju.edu.cn}
\affiliation{%
  \institution{Zhejiang University}
  \city{Hangzhou}
  \state{Zhejiang}
  \country{China}
}

\author{Liyao Li}
\authornotemark[1]
\affiliation{%
  \institution{Zhejiang University}
  \city{Hangzhou}
  \state{Zhejiang}
  \country{China}}

\author{Zhiqing Xiao}
\authornotemark[1]
\affiliation{%
 \institution{Zhejiang University}
  \city{Hangzhou}
  \state{Zhejiang}
  \country{China}}

\author{Muzhi Zhu}
\affiliation{%
  \institution{Zhejiang University}
  \city{Hangzhou}
  \state{Zhejiang}
  \country{China}}

\author{Jiaqi Hu}
\affiliation{%
  \institution{Zhejiang University}
  \city{Hangzhou}
  \state{Zhejiang}
  \country{China}}

\author{Zhanming Shen}
\affiliation{%
  \institution{Zhejiang University}
  \city{Hangzhou}
  \state{Zhejiang}
  \country{China}}

\author{Xiaomeng Hu}
\affiliation{%
  \institution{Zhejiang University}
  \city{Hangzhou}
  \state{Zhejiang}
  \country{China}}

\author{Sean Du}
\affiliation{%
  \institution{Nanyang Technological University}
  \city{Singapore}
  \country{Singapore}}

\author{Haobo Wang}
\authornote{Corresponding author.}
\email{wanghaobo@zju.edu.cn}
\affiliation{%
  \institution{Zhejiang University}
  \city{Hangzhou}
  \state{Zhejiang}
  \country{China}}

\renewcommand{\shortauthors}{Ye et al.}

\begin{abstract}
  Hallucinations in large language models (LLMs) arise from heterogeneous failure mechanisms, making reliable detection difficult for any single global uncertainty score. In this work, we formulate hallucination detection as a mechanism-aware evidence aggregation problem, where diverse representation- and token-level signals must be interpreted under multiple latent explanations. 
    We propose \textbf{FLaG}, a lightweight hallucination detection framework that models correctness through a set of latent evidence groups. Each instance is softly associated with multiple groups via an energy-based routing mechanism, and group-conditional reliability signals are combined through a principled log-marginal aggregation. This design enables FLaG to capture heterogeneous hallucination patterns while remaining invariant to decision thresholds and evaluation metrics. The framework operates as a frozen-model head, requires no modification to the underlying language model, and incurs minimal computational overhead.
    We further provide a theoretical perspective that connects FLaG to optimal evidence aggregation under heterogeneous error mechanisms, showing that the Bayes-optimal test statistic necessarily admits a log-marginal form and that FLaG constitutes a tractable approximation with a controllable error bound. Extensive experiments across multiple benchmarks and LLM backbones demonstrate that FLaG consistently achieves SOTA performance, while exhibiting robust transfer across datasets and models, and remaining effective under limited supervision.

\end{abstract}




\maketitle

\section{Introduction}

Large language models (LLMs) have demonstrated remarkable capabilities in natural language understanding and generation~\cite{zhao2023survey}. Despite this progress, LLMs are prone to hallucinations, where outputs that are fluent and seemingly coherent yet factually incorrect or unsupported. Such hallucinated responses pose serious risks in high-stakes applications, including medicine, law, and scientific decision-making, where reliability is paramount~\cite{zhang2023siren, pal2023med}. Consequently, enabling LLMs to reliably assess the truthfulness of their own generations has become a central challenge in building trustworthy LLM systems.

\begin{figure}[t]
    \centering
    \begin{subfigure}{0.236\textwidth}
        \centering
        \includegraphics[width=\textwidth]{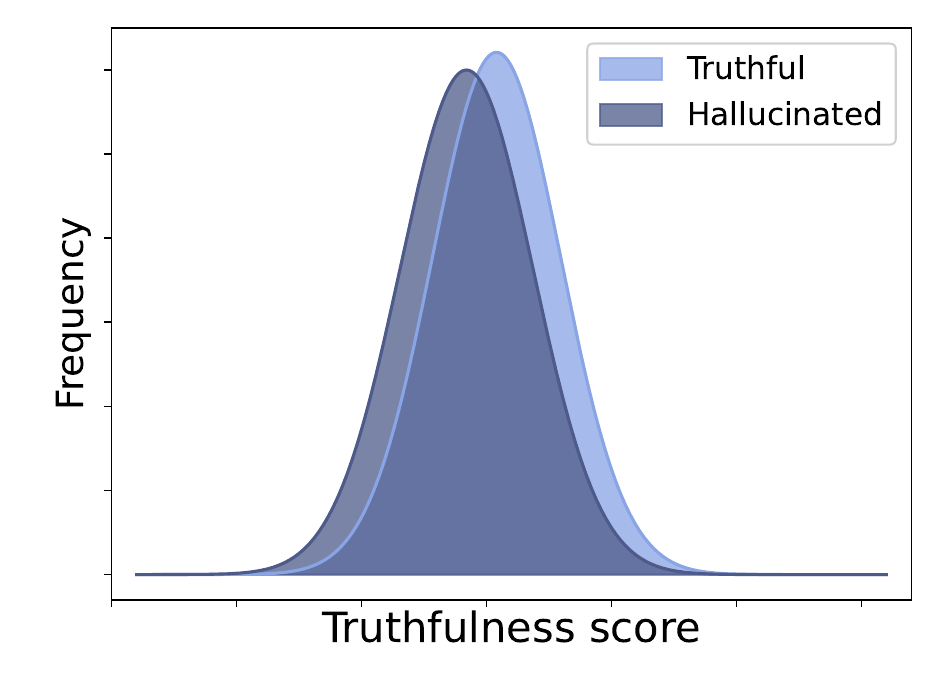}
        \caption{Vanilla Classification}
    \end{subfigure}
    \begin{subfigure}{0.236\textwidth}
        \centering
        \includegraphics[width=\textwidth]{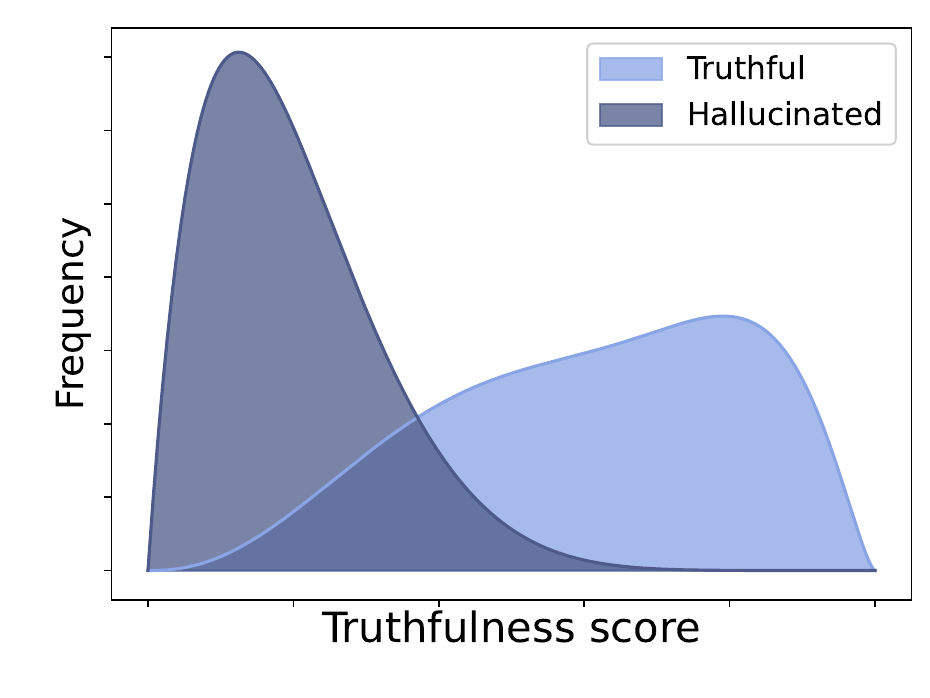}
        \caption{\ours{}}
    \end{subfigure}
    \caption{Score distributions of vanilla method v.s. \ours{}. The former is to train classifier for last-token embeddings.}
    \label{figure:intro_diff}
\end{figure}

A central challenge in hallucination detection is that hallucinations do not arise from a single, homogeneous failure mode. Empirically, hallucinated outputs exhibit diverse patterns across representation-level signals, token-level probability traces, and generation dynamics. Some failures manifest as semantic drift from the prompt, others as locally inconsistent probability assignments, and still others as overconfident yet globally implausible continuations. As a result, no single uncertainty signal or global scoring rule reliably captures all hallucination behaviors across datasets, models, and generation regimes. However, most existing detectors ~\cite{burns2022discovering, azaria2023internal, marks2023geometry, yin2024characterizing, du2024haloscope, chen2024inside, li2024inference, kossen2024semantic} implicitly assume a homogeneous notion of hallucination, collapsing all available evidence into a monolithic score. While such approaches can be effective in specific settings, they struggle to generalize (Fig. \ref{figure:intro_diff}). In practice, detectors that rely on a single view of uncertainty or a fixed decision boundary often overfit to particular hallucination types, leading to brittle performance under setting shift.

In this work, we take a different perspective. Rather than treating hallucination as a single phenomenon, we view it as the outcome of heterogeneous latent failure mechanisms. Under this view, each instance may admit multiple competing explanations of how its evidence was generated, and reliable detection requires reasoning over these alternatives. Crucially, this perspective suggests that hallucination detection is not merely a classification problem, but an evidence aggregation problem under latent mechanism uncertainty. We formalize this intuition by framing hallucination detection as the task of learning a real-valued reliability score that aggregates heterogeneous evidence sources while remaining invariant to decision thresholds and evaluation metrics. Our formulation naturally leads to a mechanism-aware scoring rule in which evidence is first interpreted under multiple latent mechanisms and then aggregated in a probabilistically coherent manner.

Based on this formulation, we propose FLaG (\textbf{F}ine-Grained \textbf{La}tent \textbf{G}rouping), a lightweight hallucination detection framework that explicitly models latent evidence groups. FLaG extracts complementary evidence from both representation geometry and probabilistic generation traces, projects them into a shared latent space, and softly associates each instance with multiple latent groups. Each group provides a group-conditional reliability signal, and these signals are combined through a principled log-evidence aggregation rule. Importantly, this aggregation does not commit to a single explanation, but marginalizes over competing latent mechanisms, allowing the detector to adaptively capture diverse hallucination patterns. FLaG is designed as a frozen-model head that requires no modification to the underlying language model and incurs minimal computational overhead. The framework naturally supports supervised, weakly supervised, and semi-supervised learning through a ranking-based objective that directly optimizes relative correctness rather than absolute labels.

Beyond empirical performance, we provide a theoretical perspective that connects FLaG to optimal evidence aggregation under heterogeneous error mechanisms. We show that, under a general mixture-of-mechanisms model, the Bayes-optimal test statistic necessarily takes a log-marginal form over mechanism-conditioned evidence. FLaG can be viewed as a tractable, learnable approximation to this optimal statistic, with an explicit approximation error bound that improves as the number of latent mechanisms increases.

Last but not least, we evaluate FLaG across multiple hallucination benchmarks and language model backbones. The results demonstrate consistent SOTA performance over strong baselines under both full and semi-supervised settings, as well as robust transfer across datasets and models. Together, these findings support the central claim of this work: hallucination detection benefits from explicitly modeling over heterogeneous latent failure mechanisms, rather than collapsing all evidence into a unified classifier.

\begin{itemize}
    \item We formulate hallucination detection as mechanism-aware evidence aggregation under heterogeneous latent failure modes, unifying representation-level and token-level signals within a single truthfulness scoring framework.
    \item We propose \textbf{FLaG}, a lightweight frozen-model detection head that softly infers latent evidence groups and combines group-conditional signals via a principled log-marginal aggregation, yielding threshold-insensitive ranking.
    \item We provide both learning and theory: a ranking-based objective that naturally extends to weakly supervised and semi-supervised settings, and a theoretical analysis connecting FLaG to the Bayes-optimal log-marginal statistic with a controllable approximation error bound.
\end{itemize}

\section{Related Work}

\subsection{Hallucination Detection}
Hallucination detection has emerged as an important research topic due to its close connection to the potential risks of deploying LLMs in real-world applications \cite{hd_survey}.
A large body of work frames hallucination detection as an uncertainty estimation problem and designs various uncertainty scoring functions.
Logit-based methods~\cite{ren2022out,malinin2020uncertainty,kuhnsemantic} directly leverage token-level probabilities to quantify uncertainty.
Verbalized methods~\cite{lin2022teaching,xiong2023can}, in contrast, prompt LLMs to explicitly produce natural language based uncertainty signals.
Consistency-based approaches~\cite{manakul2023selfcheckgpt,chen2024inside} exploit response agreement across multiple parallel samplings of the same underlying LLM.
More recent studies hypothesize that hallucination-related signals are implicitly encoded in the model's hidden states, and train classifiers to extract such signals~\cite{azaria2023internal,marks2023geometry}.
Among them, HaloScope~\cite{du2024haloscope} identifies hallucination-related subspaces via embedding decomposition,
while TSV~\cite{tsv} learns a steering vector to restructure internal representations.
The most recent work~\cite{rl4hd} formulates hallucination detection as a reasoning task and applies reinforcement learning to train LLMs to localize specific hallucinated spans from long contexts.
In contrast, \ours{} attributes hallucinations to the mixtrue of heterogeneous failure mechanisms and automatically uncovers these latent mechanisms through group-aware learning.

\subsection{Mixture-of-Experts}
Mixture-of-Experts (MoE) models have been widely studied~\cite{moe,shazeer2017outrageously} as a principled framework for modeling data heterogeneity by decomposing complex distributions into a set of specialized components.
In modern deep learning systems, MoE has been extensively adopted to improve both model capacity and efficiency, particularly in LLMs via sparse expert routing~\cite{fedus2021switch}.
Beyond serving as a scable structure, MoE has also been explored as a mechanism-aware modeling paradigm. Under this context, different experts specialize in distinct input patterns, latent factors, or failure modes~\cite{jordan1994hierarchical}.
For instance, PNs allows inputs to be softly assigned to multiple experts and aggregats expert-specific predictions~\cite{malinin2018predictive}.
In our paper, a specially designed MoE-style module is adopted as a lightweight detection framework, rather than as a generative architecture.

\begin{figure*}[htb]
    \centering
    \includegraphics[width=\textwidth]{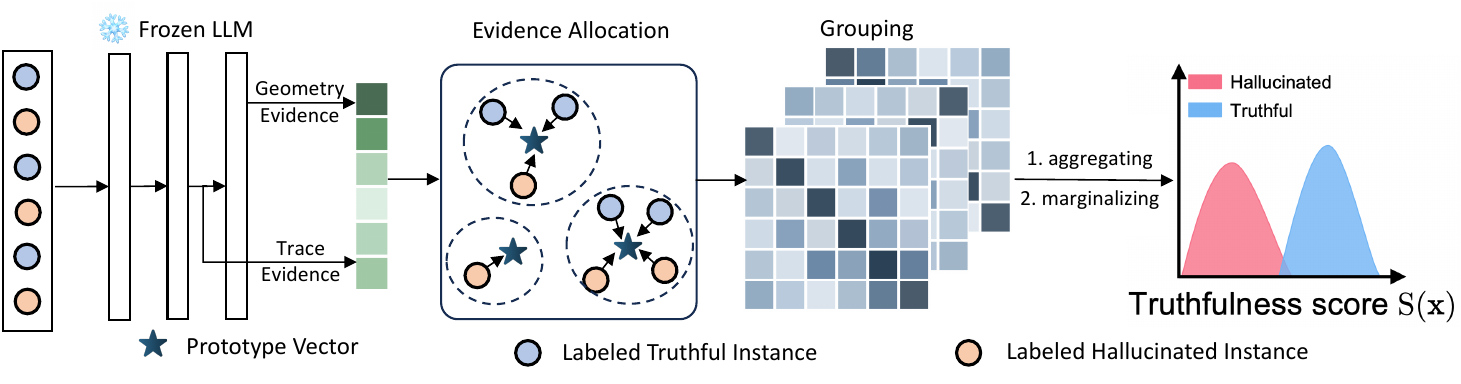}
    \caption{Overview of \textbf{FLaG}. For an instance $\mathbf{x}=(\mathbf{q},\mathbf{a})$, we extract geometry and probabilistic-trace evidence from a frozen LLM, fuse them into $\mathbf{r}(\mathbf{x})$, softly route $\mathbf{r}(\mathbf{x})$ to $K$ prototype-defined latent groups, and obtain the final truthfulness score $s(\mathbf{x})$ by log-marginal aggregation over group-wise scores..}
    \label{fig:framework}
\end{figure*}

\section{Methodology}
\subsection{Problem Formulation}
Let $\mathbf{q}=(q_1,\ldots,q_{n})$ denote a user prompt, and 
$\mathbf{a}=(a_1,\ldots,a_{m})$ denote a model-generated output.
For simplicity, we define a unified instance in detection as $\mathbf{x}=(\mathbf{q}, \mathbf{a})\in \mathcal{X}$.
The task of hallucination detection aims to identify whether the output $\mathbf{a}$ is truthful under a specified evaluation protocol.
This process can be formalized as learning a binary classifier $G:\mathcal{X}\rightarrow\{0,1\}$, where $G(\mathbf{x})=1$ indicates a truthful output, while $0$ indicates a hallucinated one.

In this paper, we consider a general learning setup in which the training set consists of a mixture of labeled and unlabeled instances.
Given a training set $\mathcal{D}$, it can be decomposed as $\mathcal{D}=\mathcal{D}_{\ell}\cup\mathcal{D}_{u}$,
where $\mathcal{D}_{\ell}=\{(\mathbf{x},y)\}$ consists of instances with supervised labels $y\in\{0,1\}$,
and $\mathcal{D}_{u}=\{\mathbf{x}\}$ consists of unlabeled instances.
The fully supervised setting corresponds to $\mathcal{D}_{u}=\varnothing$.

\subsection{Multi-View Evidence Representation}
Given an instance $\mathbf{x}=(\mathbf{q},\mathbf{a})$, our goal is to extract representations that capture evidence relevant to truthfulness (or hallucination).
In prior work, the input features to the classifier typically rely on a single global representation, e.g., the last-token embedding from the final layer \cite{tsv}.
However, the evidence of hallucination is often heterogeneous and distributed across different telemetry signals of the generation process \cite{corvushallucination}.
We therefore consider aggregating evidence from the following signal sources:
\subsubsection{Latent Geometry as Evidential Signals}
After the conditional generation $\mathbf{q}\rightarrow\mathbf{a}$, we feed the concatenated token sequence $\mathbf{q}\oplus\mathbf{a}$ back into the frozen LLM.
We then compute the mean-pooled hidden states over the prompt and output spans, denoted as $\bar{\mathbf{h}}_{\mathbf{q}}$ and $\bar{\mathbf{h}}_{\mathbf{a}}$, respectively.
The composite semantic geometry evidence is:
\begin{align}
    \boldsymbol{\psi}(\mathbf{x})
    =
    \big[
    \mathbf{h}_{\mathrm{end}} \,;\,
    \bar{\mathbf{h}}_{\mathbf{a}} \,;\,
    \bar{\mathbf{h}}_{\mathbf{a}} - \bar{\mathbf{h}}_{\mathbf{q}}
    \big]\in\mathbb{R}^{3d},
\end{align}
where $\mathbf{h}_{\mathrm{end}}$ denotes the hidden state of the last non-padding token.
We interpret these hidden states from a latent geometry view.
The $\mathbf{h}_{\mathrm{end}}$ and the pooled $\bar{\mathbf{h}}_{\mathbf{a}}$ serve as the absolute semantic coordinate of the output.
The residual $\bar{\mathbf{h}}_{\mathbf{a}}-\bar{\mathbf{h}}_{\mathbf{q}}$ measures the relative semantic drift between the output and the prompt.

\subsubsection{Probabilistic Trace as Evidential Signals}
The output $\mathbf{a}=(a_1,\ldots,a_m)$ is generated according to the conditional probability trace $\sum_{t=1}^{m}\log p(a_t \mid \mathbf{q}, a_{<t})$.
We introduce a small set of trace functionals so as to extract evidence from this trace.
The token-level log-probability is defined as $\ell_t(\mathbf{x})=\log p(a_{t}\mid \mathbf{q}, a_{<t})$, where $t\in[1,m]$ denotes the token position.
First, we compute three sample statistics, including the mean $\mu_{\ell}(\mathbf{x}) = \frac{1}{m}\sum_{t=1}^{m}\ell_t(\mathbf{x})$, the minimum $\ell_{\mathrm{min}}(\mathbf{x}) = \min_{t}\ell_t(\mathbf{x})$, and the standard deviation $\sigma_{\ell}(\mathbf{x}) = \sqrt{\frac{1}{m}\sum_{t=1}^{m}\Big(\ell_t(\mathbf{x})-\mu_{\ell}(\mathbf{x})\Big)^2}$.
Second, we compute two distributional statistics: predictive entropy and logit margin \cite{SPP}, both of which are formally token-level averages.
The former is computed as:
\begin{align}
    \mu_H(\mathbf{x}) = -\frac{1}{m}\sum_{t=1}^{m} \sum_{v\in\mathcal{V}} p(v\mid \mathbf{q}, a_{<t})\log p(v\mid \mathbf{q}, a_{<t}),
\end{align}
where $\mathcal{V}$ denotes the vocabulary.
The latter is computed as:
\begin{align}
    \mu_{\Delta}(\mathbf{x}) = \frac{1}{m}\sum_{t=1}^{m} z_{t}^{(1)} - z_{t}^{(2)},
\end{align}
where $z_{t}^{(1)}$ and $z_{t}^{(2)}$ are the largest and second-largest logits at position $t$. Notably, $z$ refers to the pre-softmax logits corresponding to the probability distribution $p$.
Third, we introduce a tail-frequency functional to count low-probability outliers:
\begin{equation}
    \rho_{\mathrm{low}}(\mathbf{x})
    =
    \frac{1}{m}\sum_{t=1}^{m}\mathbb{I}\!\left[\ell_t(\mathbf{x})<\tau_{\ell}\right],
    \label{eq:tail_freq}
\end{equation}
where $\tau_{\ell}$ is a fixed threshold. We also include the output length $m$ to serve as a complementary signal.
Collectively, the probabilistic trace evidence can be constructed as:
\begin{align}
    \boldsymbol{\phi}(\mathbf{x})
    =
    \big[
    \mu_{\ell},\ \ell_{\mathrm{min}},\ \sigma_{\ell},\ 
    \mu_H,\ \mu_{\Delta},\ \rho_{\mathrm{low}},\ m
    \big]\in\mathbb{R}^{7}.
\end{align}
This evidence captures how consistently the model assigns probability mass to the generated tokens. Moreover, the statistics in $\boldsymbol{\phi}(\mathbf{x})$ act as computable proxies of token-level likelihood ratio (App. \ref{app:theory}).

Finally, we project the two types of evidence into a shared latent space, so as to obtain a fused evidence representation:
\begin{align}
    \mathbf{r}(\mathbf{x})= f_{\mathrm{MLP}}\;\Big([f_{\mathrm{proj}}(\boldsymbol{\psi}(\mathbf{x}))\,; \boldsymbol{\phi}(\mathbf{x})]\Big)
    \in \mathbb{R}^{d},
\end{align}
where $f_{\mathrm{proj}}$ is a lightweight linear projection layer.

\subsection{Group-Aware Evidential Reasoning}
Depicted in Figure \ref{fig:framework}, we introduce $K$ latent evidence groups to adaptively discriminate heterogeneous hallucination mechanisms.

\subsubsection{Prototype-based Evidence Allocation}
For each instance $\mathbf{x}$, we allocate the extracted evidence representation $\mathbf{r}(\mathbf{x})$ to the corresponding latent group.
We represent each latent group $g \in [1,K]$ by a learnable prototype $\mathbf{c}_g \in \mathbb{R}^d$. The prototype acts as an anchor for a region in the evidence space associated with a specific hallucination mechanism.
We then define the negative energy function:
\begin{align}
    \alpha_g(\mathbf{x})
    =
    \frac{\mathbf{r}(\mathbf{x})^\top \mathbf{c}_g}
    {\|\mathbf{r}(\mathbf{x})\|_2\,\|\mathbf{c}_g\|_2}.
\end{align}
Intuitively, larger $\alpha_g(\mathbf{x})$ indicates higher compatibility between the observed evidence and mechanism $g$.
The routing distribution of groups can then be given in differentiable Boltzmann form:
\begin{align}
    \pi_g(\mathbf{x})
    =
    \frac{e^{\alpha_g(\mathbf{x})/\tau}}
    {\sum_{g'=1}^{K}e^{\alpha_{g'}(\mathbf{x})/\tau}},
    \label{eq:group_membership}
\end{align}
where $\tau>0$ controls the sharpness of the distribution.
The resulting $\pi_g(\mathbf{x})$ can be interpreted as a principled approximate posterior of $\mathbf{x}$ belonging to group $g$ in the evidence space (details deferred to App. \ref{app:theory}).
The prototypes are learned jointly with the hallucination classifier. Thus, the group structure is adaptively induced from data without requiring additional annotations.

\subsubsection{Log-Marginal Evidence Aggregation}
Next, we define how the evidence is translated into a final hallucination score.
We associate each latent group $g$ with a linear scoring function:
\begin{align}
    s_g(\mathbf{x})
    =
    \mathbf{w}_g^\top \mathbf{r}(\mathbf{x}) + b_g,
    \label{eq:group_score}
\end{align}
where $\mathbf{w}_g \in \mathbb{R}^d$ and $b_g \in \mathbb{R}$.
Different groups are trained to emphasize different dimensions of the same evidence representation, reflecting heterogeneous hallucination mechanisms.
The overall score is obtained by aggregating and marginalizing over all groups:
\begin{align}
    s(\mathbf{x})
    =
    \log \sum_{g=1}^{K}
    \pi_g(\mathbf{x}) e^{s_g(\mathbf{x})}.
    \label{eq:log_marginal_score}
\end{align}
Rather than a mixture-of-experts \cite{moe} heuristic, FLaG implements a learnable approximation to the Bayes-optimal log-evidence aggregation rule (\S~\ref{section:theory}) under heterogeneous hallucination mechanisms.
The classifier is achieved by thresholding the score: $G(\mathbf{x}) = \mathbb{I}\!\left[s(\mathbf{x}) \ge 0\right]$,
where larger $s(\mathbf{x})$ indicate higher predicted truthfulness.

\subsection{Learning Objective}
The training pipeline follows a two-stage paradigm (Algorithm \ref{alg:flag}): we fit the labeled data using a supervised objective, and then optionally incorporate unlabeled data through a semi-supervised objective.

\subsubsection{Supervised Objective.}
For labeled instances $(\mathbf{x},y)\in\mathcal{D}_{\ell}$, we adopt a margin-based ranking objective.
Given an instance pair $(\mathbf{x}^+, \mathbf{x}^-)$ with labels $y=1$ and $y=0$, respectively, we encourage the model to assign a higher score to the truthful instance:
\begin{align}
    \mathcal{L}_{\mathrm{sup}}
    =
    \mathbb{E}_{\mathbf{x}^+,\mathbf{x}^-\in\mathcal{D}_{\ell}}
    \Big[
    \log\left(1+e^{-s(\mathbf{x}^+)+s(\mathbf{x}^-)}\right)
    \Big].
    \label{eq:rank_loss}
\end{align}
This objective enforces consistent ordering between truthful and hallucinated outputs, without committing to a specific threshold.

\subsubsection{Semi-Supervised Objective.}
We further leverage unlabeled instances $\mathbf{x}\in\mathcal{D}_{u}$ to refine the preliminary latent group structure learned from supervised data.
We posit that instances strongly associated with the same group should admit a consistent relative ordering under the group-aware scoring.
Building on this intuition, we consider a subset of unlabeled instances with high routing confidence: $\mathcal{U}_g
=\big\{
\mathbf{x}\in\mathcal{D}_u
\;\big|\;
\pi_g(\mathbf{x}) \ge \gamma_g
\big\}$, where $\gamma_g$ is chosen such that $|\mathcal{U}_g|=k$.
Within $\mathcal{U}_g$, the group-wise score $s_g(\mathbf{x}$ induces a one-dimensional geometry. We form upper and lower partitions: $\mathcal{U}_g^{+} = \big\{\mathbf{x}\in\mathcal{U}_g \ \big|\ s_g(\mathbf{x}) \ge \eta_g^{+}\big\}$ and $\mathcal{U}_g^{-} = \big\{\mathbf{x}\in\mathcal{U}_g \ \big|\ s_g(\mathbf{x}) \le \eta_g^{-}\big\}$,
where $\eta_g^{+}$ and $\eta_g^{-}$ are chosen such that $|\mathcal{U}_g^{+}|=|\mathcal{U}_g^{-}|=\lfloor p|\mathcal{U}_g|\rfloor$ and $p$ is a quantile coefficient.
We then encourage the classifier $G$ to preserve the relative geometry structure within $\mathcal{U}_g^{+}$ and $\mathcal{U}_g^{-}$.
This process is enforced by a weighted group-consistent ranking loss:
\begin{equation}
    \mathcal{L}_{\mathrm{gc}}
    =
    \lambda
    \sum_{g=1}^{K} \frac{1}{K}
    \mathbb{E}_{\mathbf{x}^{+}\sim\mathcal{U}_g^{+},\ \mathbf{x}^{-}\sim\mathcal{U}_g^{-}}
    \Big[
    \log\left(1+e^{-s(\mathbf{x}^+)+s(\mathbf{x}^-)}\right)
    \Big].
    \label{eq:group_consistent_loss}
\end{equation}
Unlike explicit pseudo-labeling or full optimal transport formulations, this approach avoids constructing cost matrices or hard assignments.
To prevent noisy constraints from corrupting the supervised decision boundary, we update only $s_g(\mathbf{x})$ instead of $s(\mathbf{x})$.

Notably, both $\mathcal{L}_{\mathrm{sup}}$ and $\mathcal{L}_{\mathrm{gc}}$ operate on instance pairs sampled within each batch rather than over the entire dataset.
The overall time complexity is bounded by $O(NB)$, where $N$ denotes the dataset size and $B$ denots the batch size, thereby avoiding quadratic scaling.

\begin{algorithm}[htb]
\caption{\textbf{FLaG} Training and Inference (Fine-Grained Latent Grouping)}
\label{alg:flag}
\begin{algorithmic}[1]
\Require Frozen LLM backbone $\mathcal{M}$; labeled set $\mathcal{D}_{\ell}=\{(\mathbf{x},y)\}$; unlabeled set $\mathcal{D}_{u}=\{\mathbf{x}\}$ (optional);
number of groups $K$; temperature $\tau$; semi-supervised weight $\lambda$; group top-$k$ size $k$; quantile coefficient $p$.
\Ensure Trainable parameters $\Theta=\{f_{\mathrm{proj}}, f_{\mathrm{MLP}}, \{\mathbf{c}_g,\mathbf{w}_g,b_g\}_{g=1}^K\}$.

\State \textbf{Function} \textsc{FuseEvidence}$(\mathbf{x}=(\mathbf{q},\mathbf{a}))$
\State \hspace{1em} Extract latent geometry evidence $\boldsymbol{\psi}(\mathbf{x})$ from $\mathcal{M}$
\State \hspace{1em} Extract probabilistic trace evidence $\boldsymbol{\phi}(\mathbf{x})$
\State \hspace{1em} \Return $\mathbf{r}(\mathbf{x}) =
f_{\mathrm{MLP}}\!\left([\,
f_{\mathrm{proj}}(\boldsymbol{\psi}(\mathbf{x})) \,;\,
\boldsymbol{\phi}(\mathbf{x})
\,]\right)$

\State \textbf{Function} \textsc{Score}$(\mathbf{x})$
\State \hspace{1em} $\mathbf{r} \gets$ \textsc{FuseEvidence}$(\mathbf{x})$
\For{$g=1$ \textbf{to} $K$}
    \State $\alpha_g(\mathbf{x}) \gets
    \dfrac{\mathbf{r}^\top \mathbf{c}_g}
    {\|\mathbf{r}\|_2 \|\mathbf{c}_g\|_2}$
\EndFor
\For{$g=1$ \textbf{to} $K$}
    \State $\pi_g(\mathbf{x}) \gets
    \dfrac{\exp(\alpha_g(\mathbf{x})/\tau)}
    {\sum_{g'=1}^K \exp(\alpha_{g'}(\mathbf{x})/\tau)}$
    \State $s_g(\mathbf{x}) \gets \mathbf{w}_g^\top \mathbf{r} + b_g$
\EndFor
\State $s(\mathbf{x}) \gets
\log \sum_{g=1}^K \pi_g(\mathbf{x}) \exp\!\big(s_g(\mathbf{x})\big)$
\State \Return $s(\mathbf{x})$

\State \textbf{Stage I: Supervised training}
\For{each minibatch of labeled pairs $(\mathbf{x}^+,\mathbf{x}^-)$ with $y=1$ and $y=0$}
    \State $s^+ \gets$ \textsc{Score}$(\mathbf{x}^+)$;\quad
           $s^- \gets$ \textsc{Score}$(\mathbf{x}^-)$
    \State $\mathcal{L}_{\mathrm{sup}} \gets
    \log\!\left(1+\exp\big(-s(\mathbf{x}^+) + s(\mathbf{x}^-)\big)\right)$
    \State Update all parameters $\Theta$ using $\nabla_{\Theta}\mathcal{L}_{\mathrm{sup}}$
\EndFor

\If{$\mathcal{D}_{u} \neq \varnothing$}
\State \textbf{Stage II: Semi-supervised group-consistent refinement}
\For{each minibatch of unlabeled instances $\mathbf{x}\in\mathcal{D}_{u}$}
    \State Compute $\{\pi_g(\mathbf{x}), s_g(\mathbf{x}), s(\mathbf{x})\}_{g=1}^K$ via \textsc{Score}
    \For{$g=1$ \textbf{to} $K$}
        \State $\mathcal{U}_g \gets$ top-$k$ instances ranked by $\pi_g(\mathbf{x})$
        \State $\mathcal{U}_g^{+} \gets$ top-$p$ fraction of $\mathcal{U}_g$ ranked by $s_g(\mathbf{x})$
        \State $\mathcal{U}_g^{-} \gets$ bottom-$p$ fraction of $\mathcal{U}_g$ ranked by $s_g(\mathbf{x})$
        \State Sample $(\mathbf{x}^+,\mathbf{x}^-)$ with
        $\mathbf{x}^+ \sim \mathcal{U}_g^{+},\ \mathbf{x}^- \sim \mathcal{U}_g^{-}$
        \State Accumulate
        $\mathcal{L}_{\mathrm{gc}} \gets \mathcal{L}_{\mathrm{gc}} +
        \log\!\left(1+\exp\big(-s(\mathbf{x}^+) + s(\mathbf{x}^-)\big)\right)$
    \EndFor
    \State $\mathcal{L}_{\mathrm{gc}} \gets \dfrac{\lambda}{K}\mathcal{L}_{\mathrm{gc}}$
    \State Update only $\{\mathbf{w}_g,b_g\}_{g=1}^K$ using $\nabla \mathcal{L}_{\mathrm{gc}}$
\EndFor
\EndIf

\State \textbf{Inference:}
Given a test instance $\mathbf{x}$, output reliability score $s(\mathbf{x})$
(or prediction $G(\mathbf{x})=\mathbb{I}[s(\mathbf{x})\ge 0]$).
\end{algorithmic}
\end{algorithm}

\section{Theoretical Analysis}
\label{section:theory}
We analyze \ours{} from the perspective of composite hypothesis testing.
The proofs and further analysis for this section can be found in App. \ref{app:theory}.
 First, we assume that, conditioned on the label, the instance $\mathbf{x}$ is generated from a mixture model:
\begin{equation}
    p(\mathbf{x}\mid y)
    =
    \sum_{g=1}^{K} \pi_y(g)\, p_g(\mathbf{x}\mid y),
    \label{eq:mixture_model}
\end{equation}
where $\pi_y(g)$ denotes the label-dependent mixing weight, and $p_g(\cdot\mid y)$ denotes the group-conditional distribution.
Meanwhile, for each group , we define the group-aware log-likelihood ratio:
\begin{equation}
    \ell_g(\mathbf{x})
    =
    \log\frac{p_g(\mathbf{x}\mid y=1)}{p_g(\mathbf{x}\mid y=0)}.
    \label{eq:group_llr}
\end{equation}
\begin{theorem}
\label{thm:optimal_llr}
Under the mixture model, the Bayes-optimal log-likelihood ratio (LLR) for testing $y=1$ versus $y=0$ is given by
\begin{equation}
    \Lambda^\star(\mathbf{x})
    =
    \log \sum_{g=1}^{K}
    p(g\mid \mathbf{x}, y=0)\,
    e^{
        \ell_g(\mathbf{x})
        +
        \log \tfrac{\pi_1(g)}{\pi_0(g)}
    },
    \label{eq:optimal_llr}
\end{equation}
where
$
p(g\mid \mathbf{x}, y=0)
=
\frac{\pi_0(g)\,p_g(\mathbf{x}\mid y=0)}
{\sum_{g'}\pi_0(g')\,p_{g'}(\mathbf{x}\mid y=0)}
$
denotes the posterior distribution over latent groups under the null hypothesis $y=0$.
\end{theorem}

We therefore proves that the Bayes-optimal test statistic necessarily takes the form of a log-marginal aggregation of group-aware log-likelihood ratios, weighted by a group posterior.
Importantly, this posterior is taken with respect to the null hypothesis $y=0$, and both $p(g\mid \mathbf{x},y=0)$ and $\ell_g(\mathbf{x})$ depend on the unknown group-conditional distributions $\{p_g(\cdot\mid y)\}$.
Rather than attempting to estimate these distributions explicitly, we construct a learnable surrogate.
Specifically, \ours{} replaces the intractable posterior $p(g\mid \mathbf{x},y=0)$ with a data-driven routing distribution $\pi_g(\mathbf{x})$, and replaces the unknown group-wise log-likelihood ratio $\ell_g(\mathbf{x})$ with a parametric group score $s_g(\mathbf{x})$ computed from the fused evidence representation.
The resulting score
$
s(\mathbf{x})
=
\log \sum_{g=1}^{K}
\pi_g(\mathbf{x})\,e^{s_g(\mathbf{x})}
$
constitutes a tractable approximation to the Bayes-optimal LLR in Eq.~\eqref{eq:optimal_llr}.
Since any strictly monotone transformation of $\Lambda^\star(\mathbf{x})$ preserves the optimal ordering, this approximation suffices for reliable hallucination detection.

\subsection{Analysis of Approximation Error Bound}

We next show that \ours{} is not merely an architectural convenience: increasing the number of groups $K$ yields a provable reduction in the approximation error to the Bayes-optimal statistic.
Assume the fused evidence representation $\mathbf{r}(\mathbf{x})$ lies in a compact set $\mathcal{S}\subset\mathbb{R}^d$.
\begin{theorem}
\label{thm:approx_rate}
Assume $\Lambda^\star(\mathbf{x})$ is Lipschitz as a function of $\mathbf{r}(\mathbf{x})$ on $\mathcal{S}$.
There exists a parameter setting of \ours{} with $K$ groups such that
\begin{equation}
\sup_{\mathbf{x}}
\big|s(\mathbf{x})-\Lambda^\star(\mathbf{x})\big|
=
O\!\left(K^{-1/d}\right).
\label{eq:approx_rate}
\end{equation}
\end{theorem}

This theorem formalizes the role of fine-grained grouping. By allocating different local linear experts to different regions in the evidence space, \ours{} approximates the Bayes-optimal statistic with a controllable error that decays polynomially in $K$. Moreover, the decay exponent is determined by the evidence dimension $d$.

\begin{table*}[t]
  \centering
  \caption{
  Overall performance of baselines on the LLaMA-3-8B-Instruct, and Qwen2.5-7B-Instruct across four datasets. All results are reported in AUROC scores in percentage. The best results are in \textbf{bold} and the second best are \underline{underlined}.
  }
  \label{table:exp_overall}
  \resizebox{\linewidth}{!}{%
    \begin{tabular}{l|cccccccccc}
    \toprule
    \multirow{3.5}{*}{\textbf{Method}}
    & \multicolumn{2}{c}{TruthfulQA}
    & \multicolumn{2}{c}{TriviaQA}
    & \multicolumn{2}{c}{CoQA}
    & \multicolumn{2}{c}{TydiQA-GP}
    & \multicolumn{2}{c}{\textbf{Average}} \\
    \cmidrule(lr){2-11}
    & \makecell{LLaMA-3\\8B-Instruct} & \makecell{Qwen-2.5\\7B-Instruct}
    & \makecell{LLaMA-3\\8B-Instruct} & \makecell{Qwen-2.5\\7B-Instruct}
    & \makecell{LLaMA-3\\8B-Instruct} & \makecell{Qwen-2.5\\7B-Instruct}
    & \makecell{LLaMA-3\\8B-Instruct} & \makecell{Qwen-2.5\\7B-Instruct}
    & \makecell{LLaMA-3\\8B-Instruct} & \makecell{Qwen-2.5\\7B-Instruct} \\
    \midrule
    
    \multicolumn{11}{c}{Training-free Methods} \\
    \midrule
    Perplexity         & 61.13 & 56.40 & 74.61 & 52.68 & 62.76 & 59.86 & 52.25 & 49.13 & 62.69 & 54.52 \\
    Semantic Entropy   & 58.08 & 62.51 & \underline{79.40} & 72.79 & 55.62 & 53.40 & 55.69 & 50.21 & 62.20 & 59.73 \\
    Lexical Similarity & 51.54 & 58.33 & 77.59 & 66.32 & \textbf{79.03} & 70.12 & 62.66 & 61.98 & 67.71 & 64.19 \\
    EigenScore         & 57.12 & 53.24 & 70.16 & 69.80 & 72.59 & \underline{73.15} & \textbf{74.63} & 62.48 & 68.63 & 64.67 \\
    SelfCKGPT          & 59.66 & 64.70 & 78.77 & 70.15 & \underline{76.41} & \textbf{74.43} & 52.83 & 56.01 & 66.92 & 66.32 \\
    Verbalize          & 62.33 & 55.07 & 53.12 & 50.74 & 54.91 & 52.92 & 54.86 & 53.65 & 56.30 & 53.35 \\
    Self-evaluation    & 52.03 & 50.84 & 78.98 & 60.66 & 64.07 & 49.92 & 74.27 & 57.59 & 67.34 & 54.75 \\
    SPUQ               & 64.46 & 59.12 & 71.57 & 65.90 & 62.36 & 63.47 & 67.40 & 61.24 & 66.45 & 62.43 \\
    \midrule
    
    \multicolumn{11}{c}{Training-based Methods} \\
    \midrule
    CCS                & 52.19 & 52.90 & 60.74 & 50.17 & 50.33 & 51.89 & 72.02 & 55.44 & 58.82 & 51.57 \\
    HaloScope          & 66.79 & 69.66 & 65.31 & 64.41 & 65.83 & 61.87 & 74.08 & 66.22 & 68.00 & 65.54 \\
    Linear probe       & 70.73 & 68.01 & 74.54 & 65.07 & 69.25 & 67.28 & 70.62 & \underline{71.46} & 71.28 & 67.96 \\
    SAPLMA             & 71.68 & \underline{71.92} & 78.88 & 68.25 & 73.33 & 71.24 & 69.03 & 66.74 & \underline{73.23} & 70.04 \\
    EarlyDetec         & 66.01 & 66.15 & 69.46 & \underline{75.07} & 65.58 & 66.33 & 72.88 & 69.95 & 68.48 & 69.38 \\
    EGH                & 62.89 & 61.58 & 66.30 & 70.53 & 67.91 & 72.06 & 71.24 & 64.17 & 67.09 & 67.09 \\
    TTPD               & 68.27 & 71.10 & 71.43 & 67.54 & 70.86 & 68.77 & 69.19 & 70.63 & 69.94 & 69.51 \\
    Probe-LR           & 65.41 & 70.36 & 69.21 & 69.88 & 74.95 & 65.62 & 67.08 & 66.20 & 69.16 & 68.77 \\
    TSV                & 63.94 & 57.96 & 63.54 & 66.32 & 68.45 & 64.78 & 70.15 & 65.83 & 66.52 & 63.72 \\
    \midrule
    \rowcolor{gray!20}
    \textbf{\ours{} (Semi)} & \underline{72.49} & 70.59 & 74.63 & 74.06 & 70.20 & 69.25 & 72.27 & 67.02 & 72.40 & \underline{70.23} \\
    \rowcolor{gray!20}
    \textbf{\ours{} (Full)}   & \textbf{75.76} & \textbf{73.04} & \textbf{79.51} & \textbf{76.82} & 75.32 & 72.64 & \underline{74.41} & \textbf{72.05} & \textbf{76.25} & \textbf{73.64} \\
    \bottomrule
    \end{tabular}
    }
\end{table*}

\section{Experiments}

\subsection{Experimental Settings}

\subsubsection{Evaluation}
We conduct experiments on four benchmarks under two different settings:
(1) CoQA~\cite{reddy2019coqa} and TyDiQA-GP (English)~\cite{clark2020tydi}, where a supporting passage is provided;
and (2) TruthfulQA~\cite{lin2021truthfulqa} and TriviaQA~\cite{joshi-etal-2017-triviaqa}, where no external evidence is given.
For each dataset, we reserve 25\% of the instances as the test set, additionally sample 100 non-overlapping instances as the validation set, and use the remaining data for training.
Moreover, we consider two widely used open-source LLMs that provide accessible internal representations, namely Qwen2.5-7B-Instruct \cite{yang2024qwen2} and LLaMA3-8B-Instruct \cite{dubey2024llama}. 
By default, we used greedy sampling for the generation.
Experiments on the corresponding base models are deferred to \S \ref{subsec:overall}.
Following prior work~\cite{du2024haloscope}, to avoid sensitivity to threshold selection, we report performance using the area under the receiver operating characteristic curve (AUROC).
To obtain instance-level labels (hallucinated vs.\ truthful), we adopt an LLM-as-a-judge protocol by prompting Qwen3-235B-A22B.
We do not follow some prior studies that rely on BLEURT \cite{sellam2020bleurt}, as we observe that such metrics fail to capture fine-grained semantic discrepancies, leading to unreliable evaluation. A 100-sample human audit shows 99\% agreement for Qwen3 judge vs. 65\% for BLEURT.

\subsubsection{Baselines}
For a comprehensive comparison, we evaluate \ours{} against 17 baselines, including:
(1) self-assessment methods: Perplexity~\cite{renout}, Semantic Entropy~\cite{kuhnsemantic},
Lexical Similarity~\cite{lingenerating}, SelfCKGPT~\cite{manakul2023selfcheckgpt}, EigenScore~\cite{chen2024inside}, Verbalize~\cite{linteaching}, Self-evaluation~\cite{kadavath2022language}, and SPUQ~\cite{gao2024spuq}. (2) internal state-based methods: CCS~\cite{burns2022discovering}
, HaloScope~\cite{du2024haloscope}, 
Linear probee~\cite{pagh2007linear}, SAPLMA~\cite{azaria2023internal}, EarlyDetec~\cite{snyder2024early}, EGH~\cite{hu2024embedding}, TTPD~\cite{burger2024truth}, Probe-LR~\cite{liu2024universal}, and TSV \cite{tsv}.
Detailed configuration of each baseline is shown in App. \ref{app:baseline}.

\subsubsection{Implementation Details}
We use the frozen LLM as the backbone representation extractor.
We set the fused evidence dimension to match the backbone hidden size, and use a two-layer MLP with $1024$ hidden width for feature fusion.
Unless otherwise specified, we use $K=64$ latent groups and a fixed routing temperature of $0.1$.
All trainable components are initialized with He uniform initialization and optimized using AdamW~\cite{loshchilov2019decoupled} with a learning rate of $8 \times 10^{-4}$ and weight decay 0.01. We fix the random seed to 42 and conduct all experiments on a single NVIDIA H200 GPU using PyTorch 2.7, CUDA 12.8, and BF16 precision. 
For the supervised stage, we train for 20 epochs with a batch size of 128.
When unlabeled data are available, the semi-supervised objective is enabled after an initial warm-up of $20$ epochs.
For each group, we select the top $32$ unlabeled instances with the highest routing confidence.
Within this subset, the top and bottom $20\%$ instances ranked by the group-specific score are used to construct pseudo-ranked pairs.
The semi-supervised loss is weighted by $0.05$ relative to the supervised objective.
The unsupervised stage runs for additional $20$ epochs. Unless specified, we only discuss the semi-supervision setup in Table \ref{table:exp_overall}.
The threshold of classification during inference is set as $0.5$.
All hyperparameters are determined via empirically guided grid search on the validation set.
The global precision is set to BF16.

\subsection{Overall Performance}
\label{subsec:overall}
Table~\ref{table:exp_overall} reports the overall performance across four benchmarks and two instruction-tuned LLMs. 
We report the results of \ours{} under both the fully supervised setting (Full) and the semi-supervised setting (Semi, using only 20\% labeled instances). In contrast, training-based baselines are evaluated exclusively in the fully supervised regime.
Under full supervision, \textit{\textbf{\ours{} achieves the SOTA mean performance with a clear margin}}, yielding absolute improvements of \textbf{3.03\%} and \textbf{3.60\%} on the two LLM backbones, respectively.
Under semi-supervision, \ours{} already outperforms most fully supervised competitors.
Notably, these gains are consistent across datasets and model backbones.
Overall, the results suggest that our approach does not rely on dataset-specific shortcuts or any single uncertainty proxy; instead, it provides generalized hallucination ranking across model families and evaluation settings.

Several notable patterns emerge from the comparison. First, strong training-free methods such as Semantic Entropy, EigenScore, or lexical similarity exhibit high performance on specific datasets but fail to dominate on average. This suggests that single-view uncertainty or consistency signals are tightly coupled to particular hallucination types and answer structures. Second, although supervised probes and truthfulness classifiers (e.g., SAPLMA, Linear Probe, TSV, HaloScope) substantially improve average performance, they remain limited by a global decision boundary that implicitly assumes a homogeneous hallucination mechanism across samples. Third, the performance gap between \ours{} (Semi) and \ours{} (Full) indicates that additional supervision primarily refines the routing of samples to different evidence groups rather than merely strengthening a monolithic classifier. Overall, these observations support the hypothesis that hallucinations arise from heterogeneous generation failures, and that explicitly modeling and marginalizing over multiple latent error mechanisms enables more reliable and dataset-agnostic hallucination detection.

\begin{figure}[htb]
    \centering
    \begin{subfigure}{0.236\textwidth}
        \centering
        \includegraphics[width=\textwidth]{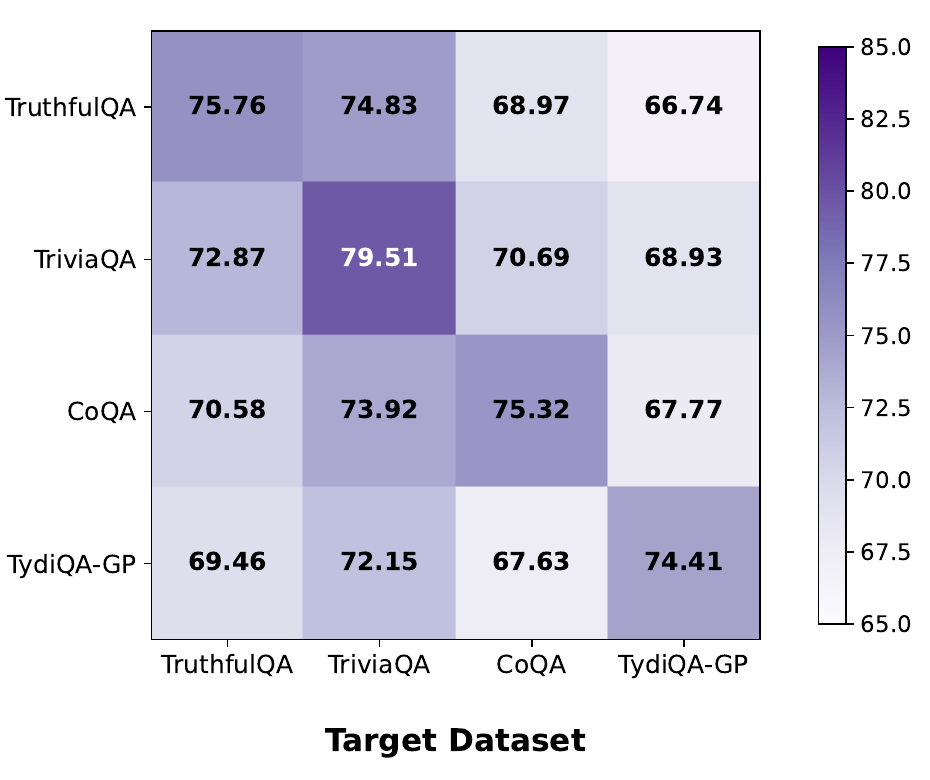}
        \caption{Dataset Transferability}
        \label{figure:exp_transfer_dataset}
    \end{subfigure}
    \begin{subfigure}{0.236\textwidth}
        \centering
        \includegraphics[width=\textwidth]{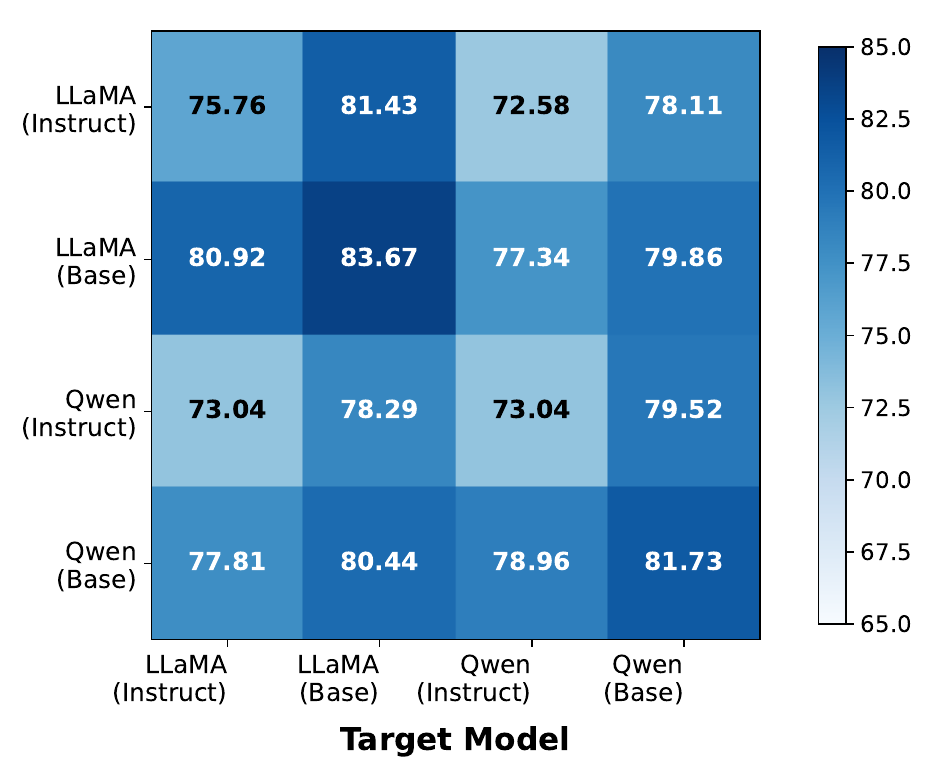}
        \caption{Model Transferability}
        \label{figure:exp_transfer_model}
    \end{subfigure}
    \caption{Transferability (reported in AUROC) across datasets and models. The vertical axis denotes the source domain or model, while the horizontal axis indicates the target.}
    \label{figure:exp_transfer}
\end{figure}

\begin{figure*}[htb]
    \centering
    \begin{subfigure}{0.245\textwidth}
        \centering
        \includegraphics[width=\textwidth]{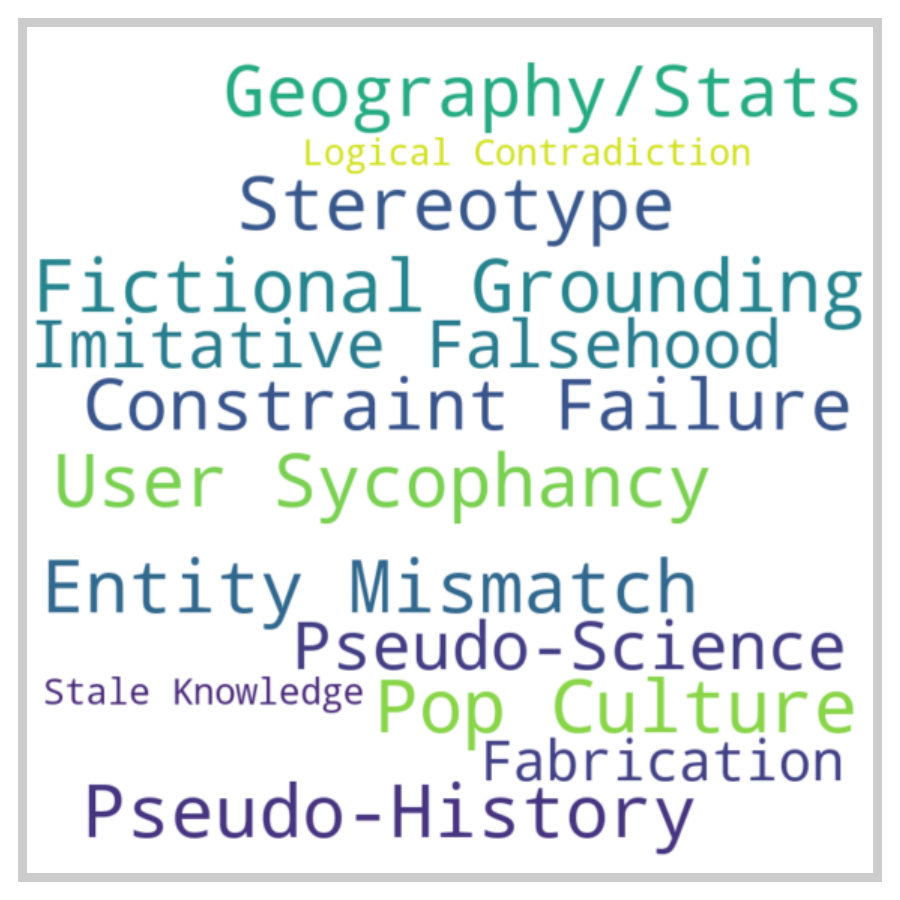}
        \caption{Top Group-1}
    \end{subfigure}
    \begin{subfigure}{0.245\textwidth}
        \centering
        \includegraphics[width=\textwidth]{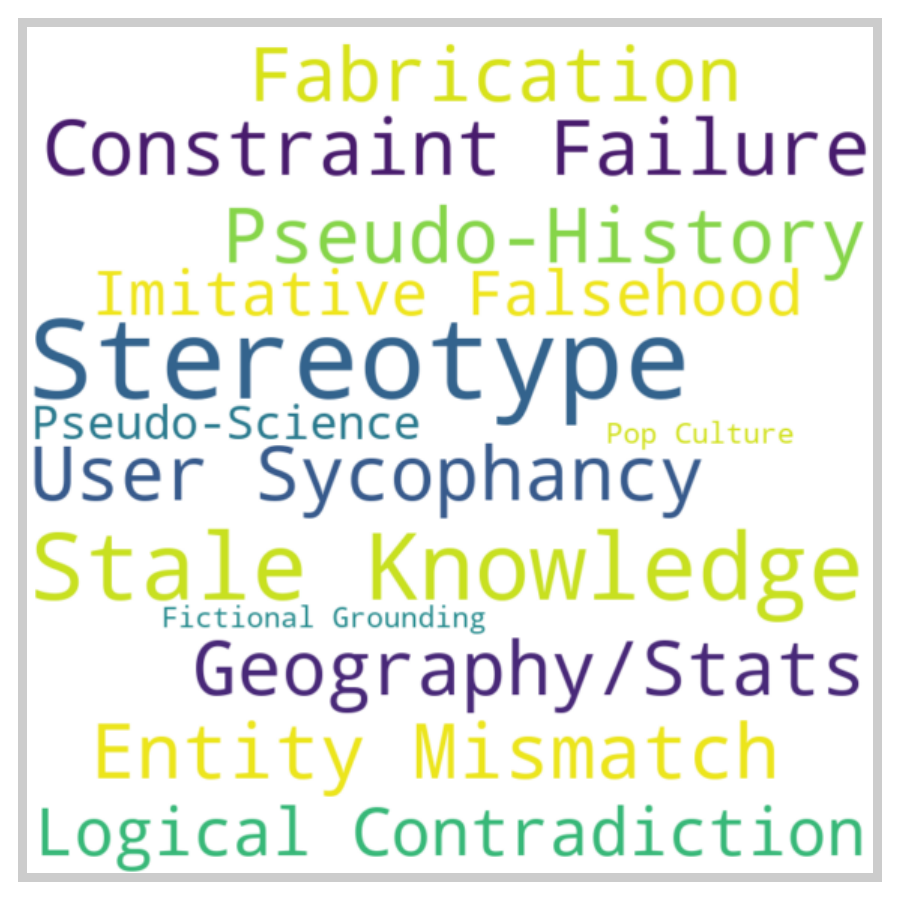}
        \caption{Top Group-2}
    \end{subfigure}
    \begin{subfigure}{0.245\textwidth}
        \centering
        \includegraphics[width=\textwidth]{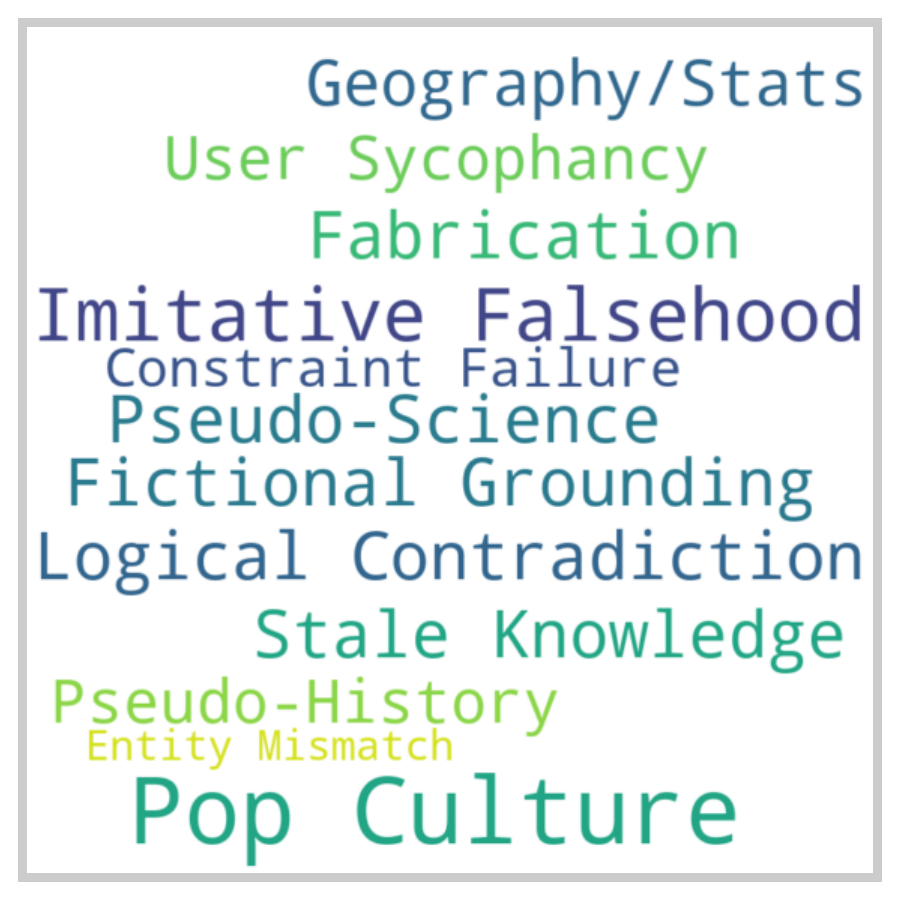}
        \caption{Top Group-3}
    \end{subfigure}
    \begin{subfigure}{0.245\textwidth}
        \centering
        \includegraphics[width=\textwidth]{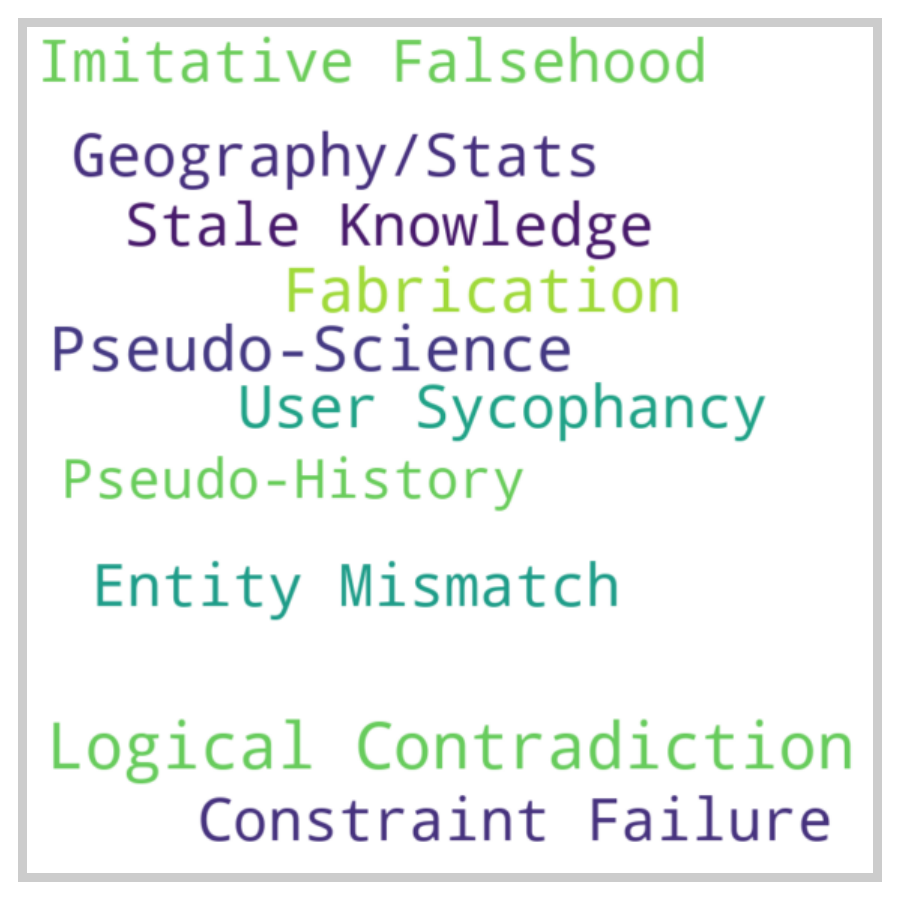}
        \caption{Top Group-4}
    \end{subfigure}
    \caption{Wordcloud interpretability of top groups. The abstract words of each groups are summarized by the Gemini3-flash.}
    \label{figure:interpretability}
\end{figure*}

\subsubsection{Transferability}
Next, we analyze the transferability of \ours{} from the perspective of test-time hallucination detection.
Under this setting, the conditions at test time (target) may differ from those used during training (source).
As shown in Figure~\ref{figure:exp_transfer_dataset}, in dataset transfer, detectors trained on a single dataset generalize well to unseen target datasets: AUROC remains consistently high and typically stays within a narrow margin of the in-domain performance. Notably, cross-dataset transfer does not collapse even when the source and target datasets differ substantially in answer format or linguistic characteristics (e.g., TriviaQA$\leftrightarrow$ TyDiQA-GP). This suggests that the learned detection signals are not tied to dataset-specific surface patterns. We also observe an expected trend that transfer is generally stronger between datasets with more similar distributions, such as TriviaQA$\leftrightarrow$ TruthfulQA, both of which are knowledge-seeking QA tasks that do not require explicit supporting passages.
Similarly, Figure~\ref{figure:exp_transfer_model} demonstrates strong cross-model transferability. Detectors trained on one backbone (e.g., LLaMA or Qwen, base or instruct variants) retain robust performance when applied to other architectures, with AUROC scores remaining above $72$ in all cross-model settings, and often exceeding $80$ when transferring from base to instruct models. This asymmetry indicates that representations learned from base models capture more generalizable hallucination-related features, whereas instruction tuning primarily introduces stylistic variability without fundamentally altering the underlying error mechanisms. Overall, these results show that our approach learns largely model- and dataset-agnostic signals of hallucination, further supporting the claim that it captures intrinsic generation failures rather than overfitting to specific domains, prompts, or architectures.

\subsubsection{Robustness}
\begin{figure}[htb]
    \centering
    \begin{subfigure}{0.236\textwidth}
        \centering
        \includegraphics[width=\textwidth]{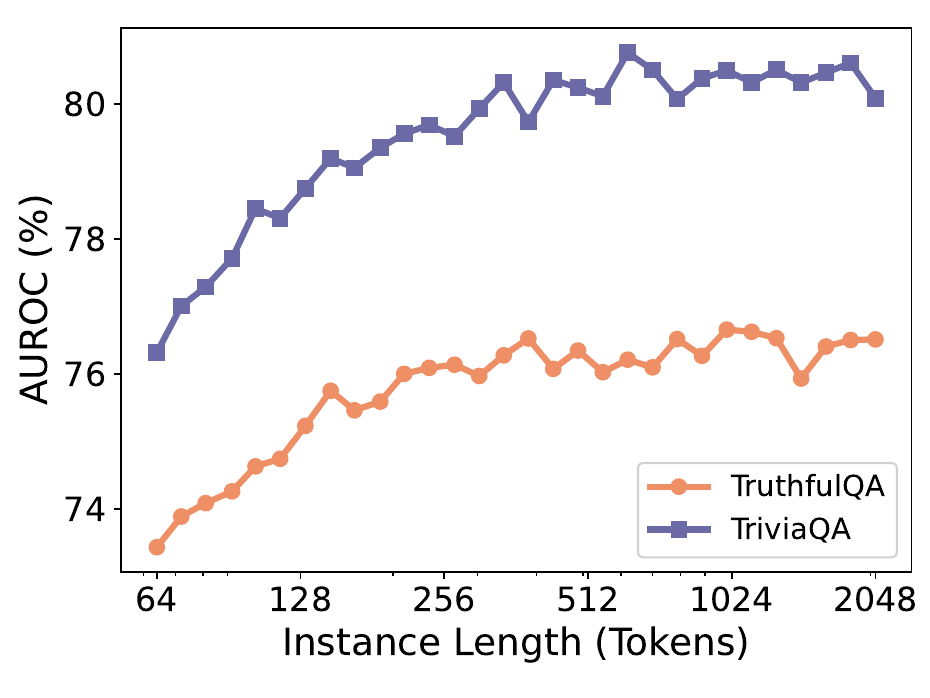}
        \caption{Complexity}
        \label{figure:exp_robustness_length}
    \end{subfigure}
    \begin{subfigure}{0.236\textwidth}
        \centering
        \includegraphics[width=\textwidth]{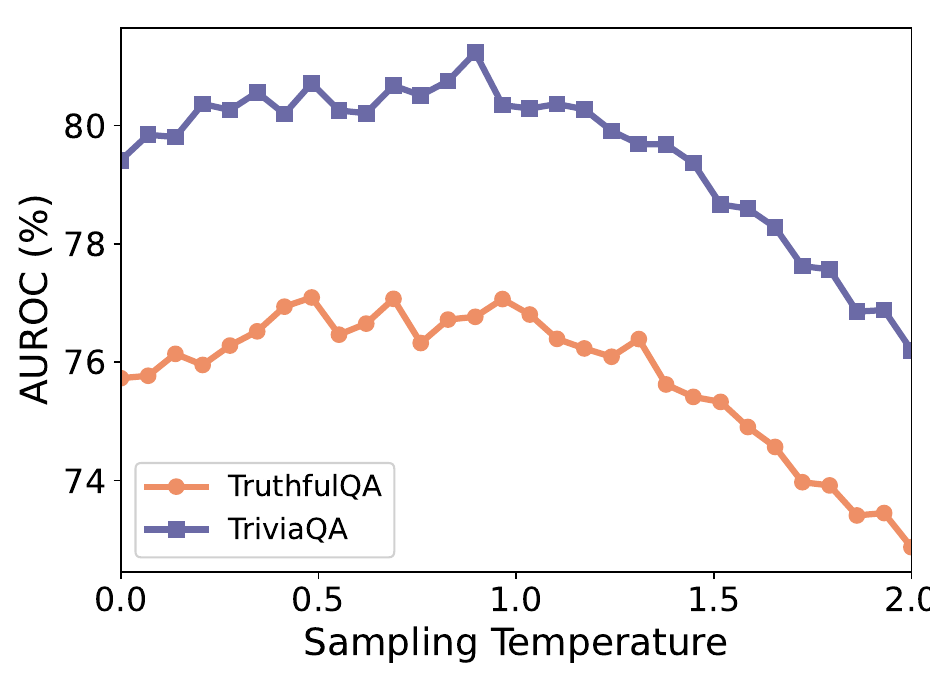}
        \caption{Diversity}
        \label{figure:exp_robustness_temperature}
    \end{subfigure}
    \caption{Robustness of instance complexity and diversity, reported on the LLaMA3-8B-Instruct and TruthfulQA.}
    \label{figure:exp_robustness}
\end{figure}

We evaluate robustness by varying instance complexity and generation diversity on LLaMA-3-8B-Instruct using TruthfulQA dataset. Instance complexity is approximated by response length, as longer generations typically involve multi-step reasoning, a larger number of entities or relations, and greater exposure to error accumulation and self-reinforcement. These elements are known to exacerbate hallucination behavior. Figure \ref{figure:exp_robustness_length} shows that detection performance consistently improves or remains stable as instance length increases, indicating that our method effectively exploits richer internal and trajectory-level signals that become more informative in complex generations, rather than being distracted by surface-level verbosity. Importantly, performance does not degrade for very long responses, suggesting resilience to compounding noise in extended outputs. Figure \ref{figure:exp_robustness_temperature} examines robustness under increasing sampling temperature, which induces higher output diversity and weaker token-level confidence. While AUROC gradually decreases at high temperatures, the degradation is smooth and limited, and performance remains competitive even when diversity substantially increases. This trend suggests that although high-temperature sampling blurs local probability cues, the method continues to rely on complementary structural and representation-level evidence. Thus, we can maintain reliable hallucination ranking across a wide range of generation regimes.

\subsubsection{Interpretability}
To provide a systematic interpretation of the latent mechanisms discovered by \ours{}, we analyze the semantic characteristics of the learned groups using an external large language model. Concretely, we use the TruthfulQA dataset as an illustrative example and assign each instance to its most activated group according to the group-wise scores produced by our model. For each group, we then collect the corresponding questions and responses and prompt Gemini3-flash to summarize their dominant error patterns in a descriptive manner. Figure~\ref{figure:interpretability} visualizes the resulting summaries using word clouds for the top four groups. The groups exhibit clear and semantically coherent distinctions: some groups are dominated by fact fabrication phenomena such as pseudo-history, stale knowledge, or entity mismatch, while others emphasize logical contradiction, constraint failure, or imitative falsehood driven by user sycophancy. Notably, these patterns align well with established taxonomies of hallucination types, despite the fact that no explicit hallucination categories are provided during training. This result suggests that the group-aware reasoning module learns to partition samples according to meaningful latent error mechanisms rather than superficial correlations, providing qualitative evidence that \ours{} captures interpretable and structured representations of hallucination behaviors.

\begin{figure*}[htb]
    \centering
    \begin{subfigure}{0.32\textwidth}
        \centering
        \includegraphics[width=\textwidth]{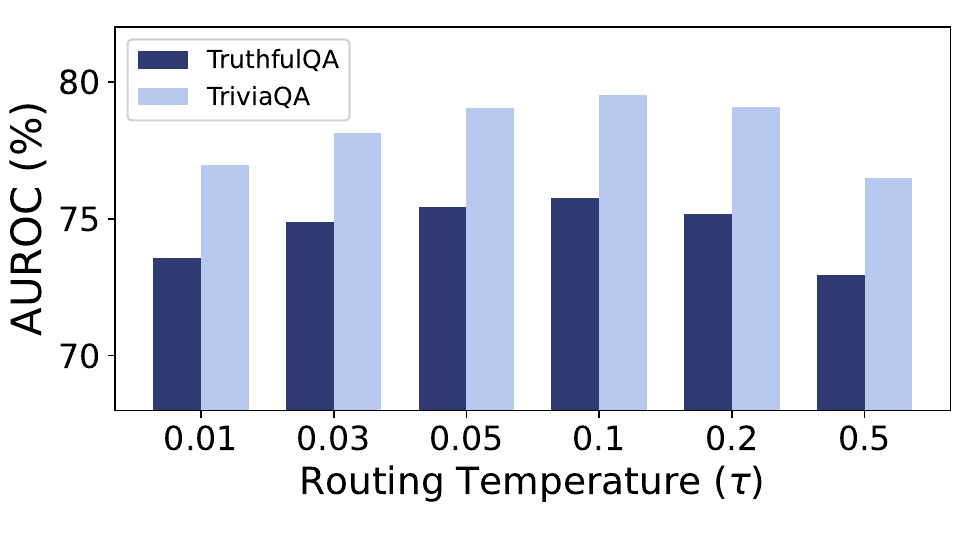}
        \caption{Distribution Level}
        \label{figure:exp_ablation_temperature}
    \end{subfigure}
    \begin{subfigure}{0.32\textwidth}
        \centering
        \includegraphics[width=\textwidth]{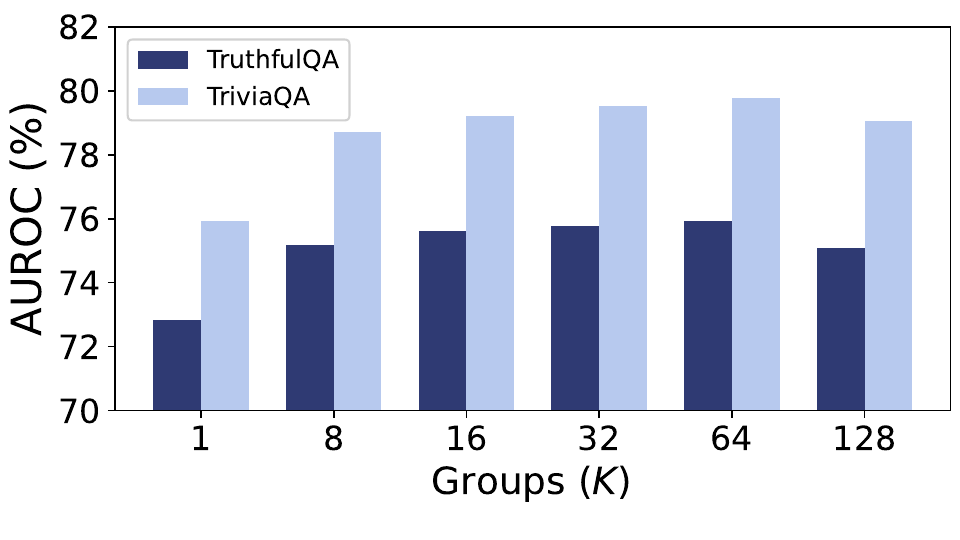}
        \caption{Group Level}
        \label{figure:exp_ablation_group}
    \end{subfigure}
    \begin{subfigure}{0.32\textwidth}
        \centering
        \includegraphics[width=\textwidth]{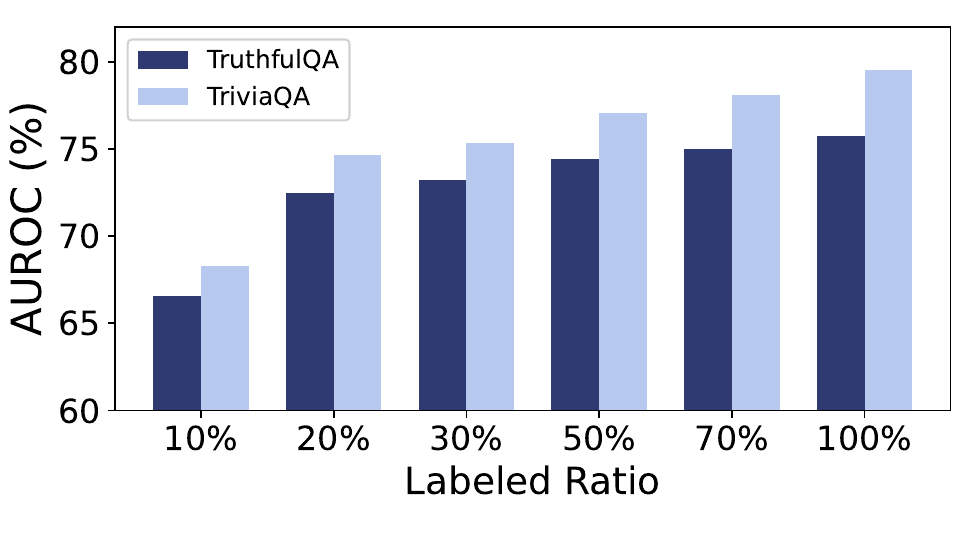}
        \caption{Supervision Level}
        \label{figure:exp_ablation_supervision}
    \end{subfigure}
    \caption{Ablation study on hyperparameters sensitivity, where the backbone LLM is based on the LLaMA3-8B-Instruct.}
    \label{figure:ablation_hyperparameters}
\end{figure*}

\subsection{Ablation Study}
We conduct ablation studies from two perspectives: the architectural components of \ours{} and its key hyperparameters.

\subsubsection{Component-Wise Ablation}
We conduct an ablation study to examine the contribution of each component in our framework, including the semantic geometric evidence representation $\boldsymbol{\psi}(\mathbf{x})$ (SGE), the probabilistic trace evidence representation $\boldsymbol{\phi}(\mathbf{x})$ (PTE), group-aware reasoning (GAR), and log-marginal aggregation (LMA). When ablating both SGE and PTE, we follow prior work and instead use the last-token hidden embedding as the input representation for classification; for GAR ablation, we reduce the number of groups to one, effectively removing mechanism-aware routing; for LMA ablation, we replace log-sum-exp aggregation with a naive weighted average. As shown in Table \ref{table:exp_ablation_component}, removing either SGE or PTE leads to a consistent performance drop across all datasets, with the degradation being more pronounced when SGE is removed, highlighting the importance of semantic geometry signals for capturing representation-level inconsistencies associated with hallucinations. When both evidence representations are removed, performance collapses sharply, indicating that simple last-token embeddings are insufficient to support reliable hallucination detection. Ablating GAR also results in a noticeable drop, suggesting that modeling heterogeneous hallucination mechanisms via group-aware routing is critical beyond simply aggregating evidence in a monolithic manner. Similarly, replacing LMA with naive averaging degrades performance, demonstrating that principled log-marginal aggregation is necessary to properly combine heterogeneous evidence sources. Overall, the full model consistently achieves the best performance across all datasets, confirming that hallucination detection benefits from the complementary interplay between rich evidence representations, mechanism-aware reasoning, and theoretically grounded aggregation.

\begin{table}[htb]
  \caption{Ablation study on varying components (reported in AUROC). The backbone LLM is the LLaMA3-8B-Instruct.}
  \label{table:exp_ablation_component}
  \centering

  \resizebox{\linewidth}{!}{%
    \begin{tabular}{cccc|cccc}
      \toprule
      \multicolumn{4}{c|}{\textbf{Component}} & \multicolumn{4}{c}{\textbf{Dataset}} \\
      \cmidrule(r){1-4}\cmidrule(l){5-8}
      \textbf{SGE} & \textbf{PTE} & \textbf{GAR} & \textbf{LMA} &
      \textbf{TQA} & \textbf{TriviaQA} & \textbf{CoQA} & \textbf{TydiQA} \\
      \midrule
      \xmark & \cmark & \cmark & \cmark & 66.31 & 70.88 & 64.27 & 63.05 \\
      \cmark & \xmark & \cmark & \cmark & 71.24 & 74.79 & 72.64 & 69.03 \\
      \xmark & \xmark & \cmark & \cmark & 55.14 & 57.02 & 54.38 & 53.61 \\
      \cmark & \cmark & \xmark & \cmark & 72.67 & 74.93 & 72.41 & 70.57 \\
      \cmark & \cmark & \cmark & \xmark & 73.84 & 77.31 & 73.58 & 72.69 \\
      \midrule
      \rowcolor{gray!12}
      \cmark & \cmark & \cmark & \cmark & \textbf{75.76} & \textbf{79.52} & \textbf{75.32} & \textbf{74.41} \\
      \bottomrule
    \end{tabular}%
  }
\end{table}

\subsubsection{Hyperparameter Sensitivity}

We analyze the sensitivity of our method with respect to key hyperparameters controlling distribution modeling, mechanism granularity, and supervision strength. Figure~\ref{figure:exp_ablation_temperature} varies the routing temperature $\tau$, which governs the sharpness of group assignment; performance remains stable over a wide range and peaks at moderate temperatures, indicating that overly sharp routing restricts evidence sharing while overly smooth routing weakens mechanism specialization. Figure~\ref{figure:exp_ablation_group} examines the number of groups $K$, where performance improves steadily as $K$ increases from small values, reflecting the benefit of modeling heterogeneous hallucination mechanisms. However, performance slightly degrades when $K$ is increased to $128$, which we attribute to insufficient data diversity to reliably support such a fine-grained partition: with limited supervision, the model lacks enough distinct evidence patterns to populate and specialize all groups, leading to underfitting and unstable group assignments. Finally, Figure~\ref{figure:exp_ablation_supervision} shows that performance improves consistently with higher labeled ratios, demonstrating that additional supervision primarily helps refine group-level specialization rather than altering the overall behavior of the model. Together, these results indicate that the proposed framework is robust to hyperparameter choices. Besides, the performance of \ours{} is maximized when the model capacity for latent mechanism modeling is well matched to the diversity and scale of the available data.

\section{Conclusions and Limitations}

We propose \textbf{FLaG}, a lightweight hallucination detector that treats hallucination detection as \emph{mechanism-aware evidence aggregation} under heterogeneous latent failure modes. 

\textbf{Limitations.}
FLaG leverages internal generation telemetry (hidden states and token-level statistics) to construct multi-view evidence; its current form is therefore not applicable in strictly black-box settings where only final text is observable. Moreover, our theoretical connection to Bayes-optimal log-evidence aggregation is established under a mixture-of-mechanisms view of evidence distributions; extending the analysis to settings with stronger distribution shift, additional conditioning signals (e.g., retrieval/tool traces), or more structured dependencies between evidence views is an important direction.

\section*{Acknowledgments}

This work is supported by the Fundamental and Interdisciplinary Disciplines Breakthrough Plan of the Ministry of Education of China. Haobo Wang is also supported by the NSFC under Grants (No. 62402424). 

\bibliographystyle{ACM-Reference-Format}
\bibliography{sample-base}

\appendix

\section{In-depth Theoretical Analysis}
\label{app:theory}

This appendix provides detailed derivations and proof steps for the theoretical statements used in the main text.
Throughout, we use $[K]=\{1,\ldots,K\}$ and denote by $\Delta_K=\{\bm{\pi}\in\mathbb{R}^K_{\ge0}:\sum_{g=1}^K\pi_g=1\}$ the probability simplex.

\subsection{Detailed Proof of Theorem~\ref{thm:optimal_llr}}
\label{app:proof_opt_llr_detailed}

Recall the mixture model (Eq.~\eqref{eq:mixture_model}):
\begin{equation}
p(\mathbf{x}\mid y)=\sum_{g=1}^K \pi_y(g)\,p_g(\mathbf{x}\mid y),
\qquad y\in\{0,1\}.
\end{equation}
Define the Bayes-optimal log-likelihood ratio (LLR)
\begin{equation}
\Lambda^\star(\mathbf{x})
=\log\frac{p(\mathbf{x}\mid y=1)}{p(\mathbf{x}\mid y=0)}.
\end{equation}
We further define the group-wise log-likelihood ratio
\begin{equation}
\ell_g(\mathbf{x})=\log\frac{p_g(\mathbf{x}\mid y=1)}{p_g(\mathbf{x}\mid y=0)},
\qquad g\in[K].
\end{equation}

\paragraph{Step 1: Expand the numerator/denominator under the mixture.}
By the mixture model,
\begin{align}
\frac{p(\mathbf{x}\mid 1)}{p(\mathbf{x}\mid 0)}
&=
\frac{\sum_{g=1}^K \pi_1(g)p_g(\mathbf{x}\mid 1)}
{\sum_{g'=1}^K \pi_0(g')p_{g'}(\mathbf{x}\mid 0)}.
\label{eq:ratio_expand}
\end{align}

\paragraph{Step 2: Multiply and divide each term by $\pi_0(g)p_g(\mathbf{x}\mid 0)$.}
For each $g$, write
\begin{equation}
\pi_1(g)p_g(\mathbf{x}\mid 1)
=
\pi_0(g)p_g(\mathbf{x}\mid 0)\cdot
\frac{\pi_1(g)}{\pi_0(g)}\cdot
\frac{p_g(\mathbf{x}\mid 1)}{p_g(\mathbf{x}\mid 0)}.
\end{equation}
Plugging into the numerator of Eq.~\eqref{eq:ratio_expand} yields
\begin{align}
\frac{p(\mathbf{x}\mid 1)}{p(\mathbf{x}\mid 0)}
&=
\frac{\sum_{g=1}^K \pi_0(g)p_g(\mathbf{x}\mid 0)\cdot
\frac{\pi_1(g)}{\pi_0(g)}\cdot
\frac{p_g(\mathbf{x}\mid 1)}{p_g(\mathbf{x}\mid 0)}}
{\sum_{g'=1}^K \pi_0(g')p_{g'}(\mathbf{x}\mid 0)}.
\label{eq:ratio_weighted}
\end{align}

\paragraph{Step 3: Identify the posterior under the null.}
Define
\begin{equation}
p(g\mid \mathbf{x},y=0)
:=
\frac{\pi_0(g)\,p_g(\mathbf{x}\mid 0)}
{\sum_{g'=1}^K \pi_0(g')\,p_{g'}(\mathbf{x}\mid 0)}.
\label{eq:null_posterior_def}
\end{equation}
Then Eq.~\eqref{eq:ratio_weighted} becomes
\begin{equation}
\frac{p(\mathbf{x}\mid 1)}{p(\mathbf{x}\mid 0)}
=
\sum_{g=1}^K
p(g\mid \mathbf{x},y=0)\cdot
\frac{\pi_1(g)}{\pi_0(g)}\cdot
\frac{p_g(\mathbf{x}\mid 1)}{p_g(\mathbf{x}\mid 0)}.
\label{eq:ratio_post_form}
\end{equation}

\paragraph{Step 4: Convert into log-sum-exp form.}
Using $\ell_g(\mathbf{x})=\log\frac{p_g(\mathbf{x}\mid 1)}{p_g(\mathbf{x}\mid 0)}$, we rewrite each multiplicative factor as an exponential:
\begin{equation}
\frac{\pi_1(g)}{\pi_0(g)}\cdot\frac{p_g(\mathbf{x}\mid 1)}{p_g(\mathbf{x}\mid 0)}
=
\exp\!\Big(\ell_g(\mathbf{x})+\log\tfrac{\pi_1(g)}{\pi_0(g)}\Big).
\end{equation}
Taking $\log$ on both sides of Eq.~\eqref{eq:ratio_post_form} yields
\begin{equation}
\Lambda^\star(\mathbf{x})
=
\log\sum_{g=1}^K
p(g\mid \mathbf{x},y=0)\,
\exp\!\Big(\ell_g(\mathbf{x})+\log\tfrac{\pi_1(g)}{\pi_0(g)}\Big),
\end{equation}
which is exactly Eq.~\eqref{eq:optimal_llr}. \hfill$\square$

\paragraph{Remark (why the posterior is under $y=0$).}
The decomposition above is obtained by factoring the denominator $\sum_{g'}\pi_0(g')p_{g'}(\mathbf{x}\mid 0)$.
If instead one factors the numerator, the posterior would be taken under $y=1$ and the residual term changes accordingly.
Both yield equivalent LLRs; we use the $y=0$ version because it aligns naturally with the ``null-posterior-weighted'' form in Eq.~\eqref{eq:optimal_llr}.

\subsection{Variational Interpretation of Prototype Routing (Detailed)}
\label{app:var_post_detailed}

We prove the variational interpretation theorem from a variational perspective with full steps.
Fix $\mathbf{x}$ and abbreviate $\mathbf{r}=\mathbf{r}(\mathbf{x})$.
Let $\alpha_g=\cos(\mathbf{r},\mathbf{c}_g)$ and define the energy $E_g=-\alpha_g$.
Consider the optimization
\begin{equation}
\min_{\bm{\pi}\in\Delta_K}\;
F(\bm{\pi})
:=
\sum_{g=1}^K \pi_g E_g
+
\tau \sum_{g=1}^K \pi_g\log\pi_g,
\qquad \tau>0.
\label{eq:var_obj_repeat}
\end{equation}

\paragraph{Step 1: Form the Lagrangian.}
Introduce a multiplier $\lambda\in\mathbb{R}$ for $\sum_g\pi_g=1$ and multipliers $\{\nu_g\ge0\}$ for $\pi_g\ge0$:
\begin{equation}
\mathcal{L}(\bm{\pi},\lambda,\bm{\nu})
=
\sum_{g=1}^K \pi_g E_g
+
\tau \sum_{g=1}^K \pi_g\log\pi_g
+
\lambda\Big(\sum_{g=1}^K\pi_g-1\Big)
-
\sum_{g=1}^K \nu_g \pi_g.
\end{equation}

\paragraph{Step 2: KKT stationarity.}
For any optimal $\bm{\pi}^\star$ with strictly positive entries (which will be implied by $\tau>0$), complementary slackness gives $\nu_g=0$.
Thus stationarity $\partial \mathcal{L}/\partial\pi_g=0$ yields
\begin{equation}
E_g+\tau(1+\log\pi_g)+\lambda=0.
\label{eq:kkt_stationary}
\end{equation}
Solving for $\pi_g$,
\begin{equation}
\log\pi_g
=
-\frac{E_g+\lambda}{\tau}-1
\quad\Longrightarrow\quad
\pi_g
=
\exp\!\Big(-\frac{E_g}{\tau}\Big)\cdot \exp\!\Big(-\frac{\lambda}{\tau}-1\Big).
\label{eq:pi_prop}
\end{equation}

\paragraph{Step 3: Enforce normalization.}
Let $Z:=\sum_{j=1}^K \exp(-E_j/\tau)$.
Summing Eq.~\eqref{eq:pi_prop} over $g$ and using $\sum_g\pi_g=1$ gives
\begin{equation}
1
=
\exp\!\Big(-\frac{\lambda}{\tau}-1\Big)\cdot Z
\quad\Longrightarrow\quad
\exp\!\Big(-\frac{\lambda}{\tau}-1\Big)=\frac{1}{Z}.
\end{equation}
Plugging back into Eq.~\eqref{eq:pi_prop} yields
\begin{equation}
\pi_g
=
\frac{\exp(-E_g/\tau)}{\sum_{j=1}^K \exp(-E_j/\tau)}
=
\frac{\exp(\alpha_g/\tau)}{\sum_{j=1}^K \exp(\alpha_j/\tau)}.
\end{equation}

\paragraph{Step 4: Uniqueness.}
The function $\bm{\pi}\mapsto \sum_g \pi_g E_g$ is linear and the negative entropy term
$\sum_g \pi_g\log\pi_g$ is \emph{strictly convex} on $\Delta_K$.
Hence $F(\bm{\pi})$ is strictly convex, implying a unique minimizer. \hfill$\square$

\paragraph{Interpretation.}
Eq.~\eqref{eq:var_obj_repeat} can be viewed as a ``free energy'':
the first term encourages assigning mass to low-energy (high-affinity) prototypes, while the second term (entropy) prevents collapse.
This provides a principled link between the prototype similarities and a posterior-like routing distribution.

\subsection{Approximation Rate: Proof Outline with Explicit Constants}
\label{app:approx_detailed}

We provide a more explicit route to the $O(K^{-1/d})$ bound in Theorem~\ref{thm:approx_rate}.
Let $\mathcal{S}\subset\mathbb{R}^d$ be compact and assume $\mathbf{r}(\mathbf{x})\in\mathcal{S}$ for all $\mathbf{x}$.
Let $\lambda^\star(\mathbf{r})$ denote the Bayes statistic expressed in evidence space, i.e.,
$\lambda^\star(\mathbf{r}(\mathbf{x}))=\Lambda^\star(\mathbf{x})$.

\begin{assumption}[Lipschitzness]
There exists $L_\lambda>0$ such that for all $\mathbf{r},\mathbf{r}'\in\mathcal{S}$,
\begin{equation}
|\lambda^\star(\mathbf{r})-\lambda^\star(\mathbf{r}')|
\le L_\lambda \|\mathbf{r}-\mathbf{r}'\|_2.
\label{eq:lipschitz_app}
\end{equation}
\end{assumption}

\paragraph{Step 1: Covering number and quantization radius.}
For $\varepsilon>0$, let $N(\varepsilon,\mathcal{S},\|\cdot\|_2)$ be the minimal number of Euclidean balls of radius $\varepsilon$ needed to cover $\mathcal{S}$.
Define the quantization radius at budget $K$:
\begin{equation}
\varepsilon_K(\mathcal{S})
:=\inf\Big\{\varepsilon>0:\; N(\varepsilon,\mathcal{S},\|\cdot\|_2)\le K\Big\}.
\end{equation}
For compact subsets of $\mathbb{R}^d$ with finite $d$-dimensional volume, there exists a constant $C_{\mathcal{S}}>0$ such that
\begin{equation}
\varepsilon_K(\mathcal{S})\le C_{\mathcal{S}} K^{-1/d}.
\label{eq:epsK_rate}
\end{equation}
(One may take $C_{\mathcal{S}}$ proportional to $\mathrm{diam}(\mathcal{S})$ and the volume ratio; standard covering arguments apply.)

\paragraph{Step 2: Construct a Voronoi partition and a piecewise-constant approximant.}
Choose centers $\{\mathbf{u}_g\}_{g=1}^K\subset\mathcal{S}$ such that
$\mathcal{S}\subset \cup_{g=1}^K B(\mathbf{u}_g,\varepsilon_K)$.
Define a Voronoi partition $\{\mathcal{S}_g\}_{g=1}^K$ by
\begin{equation}
\mathcal{S}_g
=
\Big\{\mathbf{r}\in\mathcal{S}:\;
g\in \arg\min_{j\in[K]}\|\mathbf{r}-\mathbf{u}_j\|_2
\Big\},
\end{equation}
breaking ties arbitrarily.
For any $\mathbf{r}\in\mathcal{S}_g$, we have $\|\mathbf{r}-\mathbf{u}_g\|_2\le \varepsilon_K$.
Define the approximant
\begin{equation}
\tilde{\lambda}(\mathbf{r})
=
\sum_{g=1}^K \mathbb{I}[\mathbf{r}\in\mathcal{S}_g]\;\lambda^\star(\mathbf{u}_g).
\label{eq:lambda_tilde}
\end{equation}

\paragraph{Step 3: Bound the approximation error using Lipschitzness.}
Fix $\mathbf{r}\in\mathcal{S}$ and let $g(\mathbf{r})$ be its cell index.
Then, using Eq.~\eqref{eq:lipschitz_app},
\begin{equation}
|\tilde{\lambda}(\mathbf{r})-\lambda^\star(\mathbf{r})|
=
|\lambda^\star(\mathbf{u}_{g(\mathbf{r})})-\lambda^\star(\mathbf{r})|
\le
L_\lambda \|\mathbf{u}_{g(\mathbf{r})}-\mathbf{r}\|_2
\le
L_\lambda \varepsilon_K.
\end{equation}
Taking supremum over $\mathbf{r}\in\mathcal{S}$ and using Eq.~\eqref{eq:epsK_rate} gives
\begin{equation}
\sup_{\mathbf{r}\in\mathcal{S}}
|\tilde{\lambda}(\mathbf{r})-\lambda^\star(\mathbf{r})|
\le
L_\lambda C_{\mathcal{S}} K^{-1/d}.
\label{eq:piecewise_error}
\end{equation}

\paragraph{Step 4: Realize the partition by prototype routing (soft-to-hard).}
We now connect Eq.~\eqref{eq:lambda_tilde} to the FLaG score
\begin{equation}
s(\mathbf{x})
=
\log\sum_{g=1}^K \pi_g(\mathbf{x})\,\exp(s_g(\mathbf{x})).
\end{equation}
We consider a realizable construction showing existence (as stated in Theorem~\ref{thm:approx_rate}).

\emph{(a) Prototype placement.}
Assume $\mathbf{r}$ is normalized (or we normalize within the routing).
Place prototypes $\mathbf{c}_g$ aligned with $\mathbf{u}_g$ (e.g., $\mathbf{c}_g=\mathbf{u}_g/\|\mathbf{u}_g\|_2$ if nonzero).
Then $\alpha_g(\mathbf{x})=\cos(\mathbf{r}(\mathbf{x}),\mathbf{c}_g)$ is maximized near $\mathbf{u}_g$.

\emph{(b) Low-temperature gating.}
Let $\pi_g(\mathbf{x})=\mathrm{softmax}(\alpha_g(\mathbf{x})/\tau)$ with $\tau$ small.
For any $\mathbf{x}$, let $g^\star(\mathbf{x})\in\arg\max_g \alpha_g(\mathbf{x})$.
Then for all $g$,
\begin{equation}
\pi_{g^\star}(\mathbf{x})
=
\frac{1}{1+\sum_{j\ne g^\star}\exp((\alpha_j-\alpha_{g^\star})/\tau)}
\ge
1-\sum_{j\ne g^\star}\exp\!\Big(-\frac{\alpha_{g^\star}-\alpha_j}{\tau}\Big).
\label{eq:gating_conc}
\end{equation}
If the affinity gap $\Delta(\mathbf{x}):=\min_{j\ne g^\star}(\alpha_{g^\star}-\alpha_j)>0$, then
\begin{equation}
1-\pi_{g^\star}(\mathbf{x})
\le (K-1)\exp(-\Delta(\mathbf{x})/\tau).
\label{eq:gating_tail}
\end{equation}
Thus $\pi_{g^\star}$ approaches $1$ exponentially fast as $\tau\to 0$ whenever the max is unique.

\emph{(c) Constant experts for piecewise constants.}
Set $s_g(\mathbf{x})\equiv b_g$ with $b_g=\lambda^\star(\mathbf{u}_g)$ (a special case of Eq.~\eqref{eq:group_score} by $\mathbf{w}_g=\mathbf{0}$).
Then
\begin{align}
s(\mathbf{x})
&=
\log\sum_{g=1}^K \pi_g(\mathbf{x})e^{b_g}
=
\log\Big(\pi_{g^\star}e^{b_{g^\star}}+\sum_{j\ne g^\star}\pi_j e^{b_j}\Big) \nonumber\\
&=
b_{g^\star}
+
\log\Big(\pi_{g^\star}+\sum_{j\ne g^\star}\pi_j e^{b_j-b_{g^\star}}\Big).
\label{eq:logsum_decomp}
\end{align}
Assume $|b_j-b_{g^\star}|\le B$ for all $j$ (boundedness holds on compact $\mathcal{S}$ if $\lambda^\star$ is continuous).
Then $e^{b_j-b_{g^\star}}\le e^{B}$ and Eq.~\eqref{eq:logsum_decomp} implies
\begin{align}
|s(\mathbf{x})-b_{g^\star}|
&=
\left|\log\Big(\pi_{g^\star}+\sum_{j\ne g^\star}\pi_j e^{b_j-b_{g^\star}}\Big)\right|
\le
\left|\log\Big(\pi_{g^\star}+(1-\pi_{g^\star})e^{B}\Big)\right|.
\end{align}
Using $\log(1+u)\le u$ and $\pi_{g^\star}\ge 1-(K-1)e^{-\Delta/\tau}$ from Eq.~\eqref{eq:gating_tail}, we obtain
\begin{equation}
|s(\mathbf{x})-b_{g^\star}|
\le
(1-\pi_{g^\star})e^{B}
\le
(K-1)e^{B}e^{-\Delta(\mathbf{x})/\tau}.
\label{eq:soft_to_hard_bound}
\end{equation}
Hence $s(\mathbf{x})$ approximates the hard-assignment piecewise constant $b_{g^\star}$.

\paragraph{Step 5: Combine errors.}
Let $\hat{\lambda}(\mathbf{r})$ be the hard-cell approximation induced by $g^\star(\mathbf{r})$ with values $b_g=\lambda^\star(\mathbf{u}_g)$.
Then Eq.~\eqref{eq:piecewise_error} gives
\begin{equation}
\sup_{\mathbf{r}\in\mathcal{S}}
|\hat{\lambda}(\mathbf{r})-\lambda^\star(\mathbf{r})|
\le
L_\lambda C_{\mathcal{S}}K^{-1/d}.
\end{equation}
Eq.~\eqref{eq:soft_to_hard_bound} further yields, for any $\mathbf{x}$ with affinity gap $\Delta(\mathbf{x})$,
\begin{equation}
|s(\mathbf{x})-\hat{\lambda}(\mathbf{r}(\mathbf{x}))|
\le
(K-1)e^{B}e^{-\Delta(\mathbf{x})/\tau}.
\end{equation}
Thus one may state the combined bound
\begin{equation}
\sup_{\mathbf{x}}
|s(\mathbf{x})-\Lambda^\star(\mathbf{x})|
\le
L_\lambda C_{\mathcal{S}}K^{-1/d}
+
\sup_{\mathbf{x}}(K-1)e^{B}e^{-\Delta(\mathbf{x})/\tau},
\end{equation}
which recovers the $O(K^{-1/d})$ term and makes the soft-gating residual explicit.
Allowing linear experts generally improves local approximation (smaller $B$ and tighter constants) but does not change the covering-rate exponent.

\subsection{Ranking Objective: A More Explicit Fisher-Consistency Argument}
\label{app:rank_consistency_detailed}

We expand the above theorem at the level of conditional risks.
Let $\eta(\mathbf{x})=\mathbb{P}(y=1\mid \mathbf{x})$.
Define a score function $s:\mathcal{X}\to\mathbb{R}$.
Consider the pairwise logistic loss $\varphi(u)=\log(1+e^{-u})$ and the population risk
\begin{equation}
\mathcal{R}(s)
=
\mathbb{E}_{(\mathbf{x},y),(\mathbf{x}',y')}
\Big[\,
\mathbb{I}[y=1,y'=0]\;\varphi\big(s(\mathbf{x})-s(\mathbf{x}')\big)
\,\Big],
\label{eq:pop_rank_risk_detailed}
\end{equation}
where $(\mathbf{x},y)$ and $(\mathbf{x}',y')$ are i.i.d. from the data distribution.
(Up to a constant factor, this is equivalent to sampling $\mathbf{x}^+\sim p(\cdot\mid 1)$ and $\mathbf{x}^-\sim p(\cdot\mid 0)$.)

\paragraph{Step 1: Condition on a pair $(\mathbf{x},\mathbf{x}')$.}
Let $u=s(\mathbf{x})-s(\mathbf{x}')$.
Then the conditional expected contribution of the pair is

\begin{align}
\mathcal{R}_{\mathbf{x},\mathbf{x}'}(u)
&= \mathbb{E}\!\left[\mathbb{I}[y\!=\!1,y'\!=\!0]\mid \mathbf{x},\mathbf{x}'\right]\varphi(u) \nonumber\\
&\quad + \mathbb{E}\!\left[\mathbb{I}[y\!=\!0,y'\!=\!1]\mid \mathbf{x},\mathbf{x}'\right]\varphi(-u)
\nonumber\\
&= \eta(\mathbf{x})(1\!-\!\eta(\mathbf{x}'))\varphi(u)
+ (1\!-\!\eta(\mathbf{x}))\eta(\mathbf{x}')\varphi(-u),
\label{eq:cond_risk}
\end{align}
since labels are conditionally independent given $\mathbf{x},\mathbf{x}'$.

\paragraph{Step 2: Differentiate the conditional risk.}
Using $\varphi'(u)=-\sigma(-u)$ where $\sigma(u)=\frac{1}{1+e^{-u}}$,
\begin{align}
\frac{d}{du}\mathcal{R}_{\mathbf{x},\mathbf{x}'}(u)
&=
\eta(\mathbf{x})(1-\eta(\mathbf{x}'))\varphi'(u)
-
(1-\eta(\mathbf{x}))\eta(\mathbf{x}')\varphi'(-u)
\nonumber\\
&=
-\eta(\mathbf{x})(1-\eta(\mathbf{x}'))\sigma(-u)
+
(1-\eta(\mathbf{x}))\eta(\mathbf{x}')\sigma(u).
\label{eq:cond_deriv}
\end{align}
Set the derivative to zero:
\begin{equation}
(1-\eta(\mathbf{x}))\eta(\mathbf{x}')\sigma(u)
=
\eta(\mathbf{x})(1-\eta(\mathbf{x}'))\sigma(-u).
\label{eq:stationary_eq}
\end{equation}
Using $\sigma(-u)=1-\sigma(u)$ and the identity $\frac{\sigma(u)}{\sigma(-u)}=e^{u}$, Eq.~\eqref{eq:stationary_eq} is equivalent to
\begin{equation}
e^{u}
=
\frac{\eta(\mathbf{x})(1-\eta(\mathbf{x}'))}{(1-\eta(\mathbf{x}))\eta(\mathbf{x}')}.
\end{equation}
Therefore, the unique minimizer $u^\star(\mathbf{x},\mathbf{x}')$ of $\mathcal{R}_{\mathbf{x},\mathbf{x}'}(u)$ is
\begin{equation}
u^\star(\mathbf{x},\mathbf{x}')
=
\log\frac{\eta(\mathbf{x})}{1-\eta(\mathbf{x})}
-
\log\frac{\eta(\mathbf{x}')}{1-\eta(\mathbf{x}')}.
\label{eq:u_star}
\end{equation}

\paragraph{Step 3: Implication for ordering.}
Eq.~\eqref{eq:u_star} implies that
\begin{equation}
u^\star(\mathbf{x},\mathbf{x}')>0
\quad\Longleftrightarrow\quad
\eta(\mathbf{x})>\eta(\mathbf{x}').
\end{equation}
Thus, any globally optimal scoring function must preserve the ordering of $\eta(\mathbf{x})$ almost surely (ties allowed).
In particular, taking $s^\star(\mathbf{x})=\log\frac{\eta(\mathbf{x})}{1-\eta(\mathbf{x})}$ achieves $u^\star$ for every pair.
More generally, any strictly increasing transform of $\eta(\mathbf{x})$ yields the same ordering. \hfill$\square$

\section{Dataset Specifications}
\label{app:datasets}

We use two prompt templates according to whether the dataset provides
supporting context. For context-free QA datasets, including TruthfulQA and
TriviaQA, the prompt is:
\begin{quote}
\small\texttt{Answer the question concisely. Q: <question> A:}
\end{quote}
For context-dependent datasets, including TyDiQA-GP and CoQA, the prompt is:
\begin{quote}
\small\texttt{Answer these questions concisely based on the context: \textbackslash n Context: <passage context> Q: <question> A:}
\end{quote}

\section{Baseline Implementation Details}
\label{app:baseline}

For Perplexity~\cite{renout}, we use the official implementation and average
perplexity over generated tokens. For sampling-based baselines, we follow the
original configurations and generate 10 samples with temperature 0.5. Lexical
Similarity~\cite{lingenerating} uses ROUGE-L; SelfCKGPT~\cite{manakul2023selfcheckgpt}
uses the recommended NLI variant with a fine-tuned DeBERTa-v3-large model;
HaloScope~\cite{du2024haloscope} and EGH~\cite{hu2024embedding} use their
official or released codebases.

For Verbalize~\cite{linteaching}, we use the following confidence-elicitation
prompt:
\begin{quote}
\small\texttt{[Context: <context>] Q: <question> A: <answer>. \textbackslash n The proposed answer is true with a confidence value (0-100) of,}
\end{quote}
where the context field is omitted for context-free datasets. The generated
confidence value is directly used as the uncertainty score.

For Self-evaluation~\cite{kadavath2022language}, we use:
\begin{quote}
\small\texttt{[Context: <context>] Question: <question> \textbackslash n Proposed Answer: <answer> \textbackslash n Is the proposed answer: \textbackslash n (A) True \textbackslash n (B) False \textbackslash n The proposed answer is:}
\end{quote}
Again, the context field is omitted when unavailable. Following the original
paper, we use the log probability of token ``A'' as the uncertainty score.

\section{LLM Usage Statement}
LLMs are used in this work solely for language polishing and
presentation purposes.
Specifically, LLMs are employed to improve clarity, grammar, and readability of the
manuscript text written by the authors.
They are \emph{not} used for designing the proposed method, generating experimental
results, selecting hyperparameters, analyzing outcomes, or drawing scientific
conclusions.
All technical content, experimental design, and empirical findings are entirely
produced and verified by the authors.

\end{document}